\documentclass[11pt]{article}
\usepackage[]{acl}
\usepackage{microtype}
\usepackage{enumitem}
\usepackage[OT1]{fontenc}
\usepackage{times}
\usepackage{booktabs}
\usepackage{multirow}
\usepackage{xspace}
\usepackage{adjustbox}
\usepackage{inconsolata} 
\usepackage[capitalise,noabbrev]{cleveref}
\usepackage[boldmath]{numprint}
\usepackage{amssymb}
\usepackage{pifont}
\usepackage{subfig}

\definecolor{light-gray}{gray}{0.95} 
\definecolor{very-light-gray}{gray}{0.98} 

\crefname{appendix}{Appendix}{Appendices}
\Crefname{appendix}{Appendix}{Appendices}

\newcommand{\BhashaAbhijnaanam}{B-A\xspace}
\newcommand{\ngram}{$n$-gram\xspace}
\newcommand{\ngrams}{$n$-grams\xspace}

\newcommand{\removed}[1]{}
\newcommand{\REMOVED}[1]{}

\newcommand{\xmark}{\ding{55}}%

\title{Improving Informally Romanized Language Identification}

\author{Adrian Benton,$^\dag$ Alexander Gutkin,$^\circ$ Christo Kirov,$^\dag$ Brian Roark$^\ddag$ \\
  Google Research, $^\dag$New York; $^\circ$London, UK; $^\ddag$Portland, OR \\
  \texttt{\{adbenton,agutkin,ckirov,roark\}@google.com} \\
  }

\begin{document}
\maketitle

\begin{abstract}
The Latin script is often used to informally write languages with non-Latin native scripts. In many cases (e.g., most languages in India), the lack of conventional spelling in the Latin script results in high spelling variability.
Such romanization renders languages that are normally easily distinguished due to being written in different scripts -- Hindi and Urdu, for example --  highly confusable.
In this work, we increase language identification (LID) accuracy for romanized text by improving the methods used to synthesize training sets. We find that training on synthetic samples which incorporate natural spelling variation yields higher LID system accuracy than including available naturally occurring examples in the training set, or even training higher capacity models. 
We demonstrate new state-of-the-art LID performance on romanized text from 20 Indic languages in the Bhasha-Abhijnaanam evaluation set~\cite{madhani-etal-2023-bhasa}, improving test F1 from the reported 74.7\% (using a pretrained neural model) to 85.4\% using a linear classifier trained solely on synthetic data and 88.2\% when also training on available harvested text.
\end{abstract}

\section{Introduction}

Web crawls are an important source of multilingual training data for natural language modeling, containing rich and diverse spontaneous language, albeit alongside less useful content such as boilerplate and non-language text. 
Due to this latter noise, to reach adequate levels of data quality, 
aggressive filtering is typically applied, including setting thresholds on language identification (LID) confidence to retain text \cite[see, e.g.,][]{raffel:2020,xue-etal-2021-mt5,abadji-etal-2022-towards,kudugunta:2023,kargaran2024glotcc}.

Filtering text by LID confidence places languages that are less confidently identified at a higher risk of being filtered out, and, indeed, certain classes of text are heavily underrepresented in such collections as a result, such as the romanized text detailed below.
This can become a self-reinforcing status, to the extent that such collections may be used as sources of training data for subsequent LID systems.  For example, informal romanization -- the use of the Latin script without orthography to write languages that are natively written with non-Latin scripts -- is both common, e.g., in the languages of South Asia, and generally makes the languages much more difficult to identify. The 22 scheduled languages of India all use non-Latin scripts in their official writing systems, but are also commonly written informally in the Latin script~\cite{brandt:2020}. Even so, romanized texts in these languages are mostly omitted from large, multilingual web corpora such as multilingual C4 (mC4,~\citealt{xue-etal-2021-mt5}),\footnote{mC4 contains romanized text in 6 languages that natively use other scripts, but only one South Asian language was included (Hindi), and quality assessments of the romanized sets in mC4 have been unfavorable \citep{kreutzer-etal-2022-quality}.} and are relatively sparse in MADLAD-400~\cite{kudugunta:2023}\footnote{\url{https://huggingface.co/datasets/allenai/MADLAD-400}, data released under CC-BY-4.0 license.} and GlotCC~\cite{kargaran2024glotcc}.\footnote{\url{https://github.com/cisnlp/GlotCC}, data released under CC0-1.0 license. See \cref{tab:natural_counts} for data sizes.}

The distinction between LID system performance on text in the native scripts of languages versus romanized text in those languages is strikingly demonstrated in the Bhasha-Abhijnaanam (denoted \BhashaAbhijnaanam) LID benchmark~\cite{madhani-etal-2023-bhasa}\footnote{\url{https://huggingface.co/datasets/ai4bharat/Bhasha-Abhijnaanam}, data released under CC0 license.} for the 22 scheduled languages of India. The best systems reported in that paper achieved nearly 99\% accuracy on the native script LID task, while the best reported performance on the romanized task (covering 20 of the 22 languages) reached just over 80\% accuracy.  Further, to achieve this latter result, they relied on a relatively expensive BERT model~\citep{devlin-etal-2019-bert}; a simple fastText~\cite{joulin2016fasttext,joulin-etal-2017-bag} linear model yielded the best performance on native script LID, but on the romanized task its performance fell far behind the best system at just 71.5\% accuracy.

Given the dearth of natural, collected training data in the Latin script for these languages, the above romanized LID results were achieved using synthesized training data created by automatically romanizing native script training sets using a neural sequence-to-sequence transliteration model \cite{madhani-etal-2023-aksharantar}. In this paper, we evaluate methods for improving romanized LID system performance, including synthesizing training data in a way that mimics natural spelling variation in romanized text; and augmenting synthesized data with distantly supervised datasets (MADLAD-400 and GlotCC). 
Through careful controlled experimentation, we demonstrate that: 
\begin{itemize}[noitemsep,topsep=0pt,parsep=0pt,partopsep=0pt]
\item Synthesizing training data with spelling variation by sampling from romanization models provides very large accuracy improvements for this task, even when using smaller, less accurate romanization models;
\item These improvements are obtained to a larger extent by lightweight linear models, so that higher capacity pretrained models are not needed to reach the best reported results; and
\item Augmenting the training data by including diversely synthesized copies of the training set and/or some independently harvested data yields further modest improvements.
\end{itemize}
Our best system reduces the best previously reported error rate on this task by over 60\% relative.\footnote{Explicit system details beyond those provided in this paper along with any new resources and scripts required to replicate these results are available at \url{https://github.com/google-research/google-research/tree/master/informally_romanized_lang_id}.\label{footnote:url}}

We also provide extensive analysis of language confusions, system errors and the kinds of spelling variation produced by our synthesis methods.

\section{Background and preliminaries}
\subsection{Language identification (LID) systems}
LID is a task with a long history, dating from at least \citet{mustonen:1965}.
\citet{jauhiainen2019automatic}'s survey notes that LID is typically cast as a text classification task over a closed set of mutually exclusive classes.\footnote{Multilingual text, which would seem to contradict this idea of mutual exclusivity, is composed of shorter monolingual text spans for which a single tag per span would suffice.}  They also note the early adoption of character \ngrams as features in diverse LID classification approaches \cite{church1986stress,beesley1988language,cavnar1994n}, which remains a key feature set for modern LID~\cite{lui-baldwin-2012-langid,brown-2014-non,kargaran-etal-2023-glotlid,burchell-etal-2023-open} and dialect identification systems~\cite{coltekin-etal-2018-tubingen,baimukan-etal-2022-hierarchical}.

Document level LID assigns labels to entire documents, within which there is typically redundant evidence, hence the 
classification task is easier.
Shorter samples, including sub-sentential strings, present more of a challenge, but such systems can be applied in more scenarios than document-level systems, including, e.g., identifying different language spans in multilingual documents.

As stated in the introduction, LID systems are often used for analyzing or filtering large text collections, hence efficient inference is a key consideration.  Generally speaking, feature-based linear classifiers are practical alternatives to more computationally expensive neural methods~\cite{toftrup-etal-2021-reproduction,adebara-etal-2022-afrolid}, including those that might be based on pretrained language models~\cite{mohan:2023,manukonda-kodali-2025-bytesizedllm}. For challenging scenarios, such as when two languages are highly confusable, more expressive classifiers can yield accuracy improvements that justify the extra computational expense.

While modern neural language models have been shown to perform well on general LID tasks across a range of world languages, their performance depends on the subset of languages the model is prompted to identify and the scripts these languages are written in \cite{chen-etal-2024-fumbling}. More importantly, such models are extremely compute and memory-intensive to be deployed at scale, e.g., at the corpus filtering stage. In fact, \citet{kargaran-etal-2023-glotlid} explicitly cites inference throughput as a consideration for developing the GlotLID language identification system, which was used to filter the GlotCC corpus mentioned earlier.

\subsection{LID of romanized language}\label{sec:romanized_lid}



LID systems based on lightweight linear models typically rely on features consisting of short collocations --- character \ngrams{} --- to distinguish between languages. When two languages use different scripts, which are encoded in distinct Unicode blocks, even character unigrams can be sufficient to distinguish the two.  Written in their native scripts, Chinese and English are straightforward to distinguish, hence are unlikely to be confused by a classifier; similarly Hindi and Urdu -- the former written in the Devanagari script, the latter in Perso-Arabic -- are quite easy to distinguish given the lack of character overlap.  However, when Hindi and Urdu are written in the Latin script (romanized), not only do they share the same characters, but the character collocations observed in the two highly mutually-intelligible languages will greatly overlap. For example, \cref{tab:hi_romanization_examples} presents an example of a word shared by Hindi and Urdu with multiple overlapping attested romanizations in both languages. 

\begin{table}[t]
  \centering%
  \includegraphics[width=0.8\linewidth]{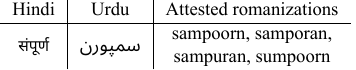}\vspace*{-0.05in}
  \caption{\footnotesize Human-attested romanizations of a word in Hindi and Urdu that translates as `complete'' in English. These attested romanizations are shared across both Hindi and Urdu.}\vspace*{-0.2in}
  \label{tab:hi_romanization_examples}
\end{table}


Accurate LID of romanized text is of practical importance as the Latin script is often used as an informal medium between speakers of languages that are formally written in a different script, e.g., in text messages or social media posts~\cite{ahmed-etal-2011-challenges,irvine-etal-2012-processing,adouane-etal-2016-romanized,baruah2024transliteration,perera-etal-2025-indonlp}. The lack of (or limited access to) standard spelling conventions for these languages in the Latin script
is due to multiple confounding sociolinguistic, technological and political factors at play in South Asia~\cite{choksi:2020,brandt:2020} and beyond~\cite{bahri:2022}, and poses yet another challenge for romanized LID. Additional practical uses of romanized LID include improving romanized text training data coverage for modern neural language models~\cite{caswell-etal-2020-language}, LID of romanized text entry in mobile keyboards~\cite{wolf-sonkin-etal-2019-latin}, extending neural methods to poorly documented languages~\cite{post:2017,aji-etal-2022-one} or languages without orthography~\cite{torwali:2020}, among others.

Several datasets have been released that contain parallel native/Latin script data, which permits, among other things, training and/or validation of transliteration models (see \cref{sec:translit_background}). These include the Dakshina dataset~\cite{roark-etal-2020-processing}, which includes single word native/Latin script pairs in romanization lexicons as well as native/Latin script parallel full sentences, in addition to native-script-only Wikipedia text, for twelve South Asian languages. The Aksharantar dataset~\cite{madhani-etal-2023-bhasa,madhani-etal-2023-aksharantar} contains mined single word native/Latin script pairs in romanization lexicons for 21 South Asian languages. And, of course, the benchmark being used for this work, the \BhashaAbhijnaanam dataset, also contains parallel resources of this sort.

Literature on romanized LID \textit{per se} is relatively scarce. In one of the earliest works on the subject, \citet{pavan:2010} propose a string similarity-based system for identification of romanized documents in Hindi, Telugu, Tamil, Kannada and Malayalam. More recently, \citet{nielsen-etal-2023-distinguishing} proposed approaches for distinguishing romanized Hindi from romanized Urdu in the Dakshina dataset. \citet{dey:2024} compared support vector machines (SVM,~\citealp{cortes:1995}) and finetuned XLM-RoBERTa architecture~\cite{conneau-etal-2020-unsupervised} for romanized LID using data from 12 South Asian languages sourced from the \BhashaAbhijnaanam and Aksharantar datasets. Beyond South Asia, \citet{adouane-etal-2016-romanized} investigated LID of romanized Arabic and Berber using SVM classifiers trained on word- and character-level \ngram features.

\subsection{Classifiers}\label{sec:classifiers}
We consider two classes of LID models in this paper: fastText linear models~\cite{joulin2016fasttext,joulin-etal-2017-bag}\footnote{\url{https://fasttext.cc/}, MIT license.} and neural pretrained transformer-based multilingual T5 (mT5) models~\cite{xue-etal-2021-mt5}.\footnote{\url{https://github.com/google-research/multilingual-t5}, Apache License 2.0.} 

While linear classifiers have low inference latency, pretrained mT5 classifiers are more expressive, and may be able to better disambiguate similarly written languages. Since prior results demonstrated that pretrained neural models outperformed linear models~\cite{madhani-etal-2023-bhasa}, 
we use mT5 to gauge whether lightweight methods can provide competitive results to such models, and to investigate the interaction between training set and classifier capacity.\footnote{Also, mT5 pretraining is unlikely to be contaminated with the training or validation data for the current task, since the model predates the task. Additionally, unlike modern LLMs, it has not been instruction-tuned. For these reasons, it provides an informative comparison within our controlled experiments.} See \cref{app:langid_training} for training details.

\subsection{Transliteration models}\label{sec:translit_background}
Transliteration of text from one script to another is typically framed as a sequence-to-sequence task. While this task can be performed with neural sequence-to-sequence models, non-neural methods are also competitive. \citet{kirov-etal-2024-context} recently examined romanization (transliteration from native to Latin script) methods for synthesizing full sentence parallel text, and evaluated the character error rate (CER) achieved by multiple model types.  For single, non-ensembled models producing single best romanizations, CER\footnote{CER was assessed against the closest match from multiple references, hence is a minimum CER over that set.} ranged from 2.6\% from a fully pretrained T5 model to 3.7\% for a non-neural pair \ngram method; these results were slightly improved via further system ensembling.
A more complex evaluation of $k$-best outputs showed a similar pattern of relatively narrow ranges of error rates.
In this paper, we use non-neural pair \ngram methods for automatic romanization, trained following the approach outlined in \citet{kirov-etal-2024-context}.

Briefly, pair \ngram models \cite{bisani2008joint} are a class of models used to map between two sets of discrete token sequences, and are commonly used for transliteration \cite{hellsten-etal-2017-transliterated,kirov-etal-2024-context}. They can be directly encoded as weighted finite-state transducers (WFSTs), which permit efficient exact inference.  Given a parallel corpus of input/output strings, 
expectation maximization (EM) is used to create strings of aligned single character input/output pairs, from which \ngram models are trained.  For example, if {\tt ABC} is aligned with {\tt abc}, then the EM alignment algorithm may produce {\tt A:a B:b C:c}, and this string (along with a full corpus of other such alignments) are used to train an \ngram model.  The pair symbols are then split into input and output symbols, so that the \ngram model is encoded as a transducer.  See \citet{kirov-etal-2024-context} for details on training such models from romanization lexicons in the Dakshina dataset~\cite{roark-etal-2020-processing}, which we also use.\footnote{\url{https://github.com/google-research-datasets/dakshina}, data released under CC BY-SA 4.0 license.}

The LID benchmark that we investigate in this paper provided romanized training and development data synthesized using IndicXlit \cite{madhani-etal-2023-aksharantar}, a neural sequence-to-sequence transliteration model trained on over twenty south Asian languages from the Aksharantar romanization lexicon, a lexicon we also use in this paper.\footnote{\url{https://huggingface.co/datasets/ai4bharat/Aksharantar}, data released under CC-BY and CC0 licenses.} 


\section{Methods}

\subsection{Evaluation}
We evaluate classifiers on the \BhashaAbhijnaanam romanized LID task~\cite{madhani-etal-2023-bhasa}. 
The benchmark provides training/development data for each language in the native script, and the IndicXlit system, described in \cref{sec:translit_background}, was used to romanize the native script text to also provide synthetic training/development data in the Latin script for 20 of the 22 languages.


The \BhashaAbhijnaanam romanized test set consists of full sentence examples from Dakshina~\cite{roark-etal-2020-processing} for the following 11 languages: Bangla, Gujarati, Hindi, Kannada, Malayalam, Marathi, Punjabi, Sindhi, Tamil, Telugu, and Urdu. Each of these languages contains between 4,371 to 4,881 test examples. The additional languages in the romanized test set
(Assamese, Bodo, Kashmiri, Konkani, Maithili, Manipuri, Nepali, Oriya, and Sanskrit)\footnote{The romanized benchmark omits Dogri and Santali.} each contain roughly 10\% as many test examples as the Dakshina set (423 to 512 examples per class).


For all runs, languages in the training set are oversampled to the plurality class -- the language with the most examples in a given training set -- to ensure a uniform prior distribution over languages. For instance, if our training set contains 1,000 Hindi, 750 Bangla, and 500 Bodo examples, we oversample the Bangla and Bodo examples to yield 3,000 total training examples, by duplicating 250 Bangla and 500 Bodo examples. We train fastText models from scratch on the various training sets and also use these training sets to finetune the pretrained public base and large mT5 checkpoints.

\begin{table*}[t]
    \centering%
    \includegraphics[width=0.8\linewidth]{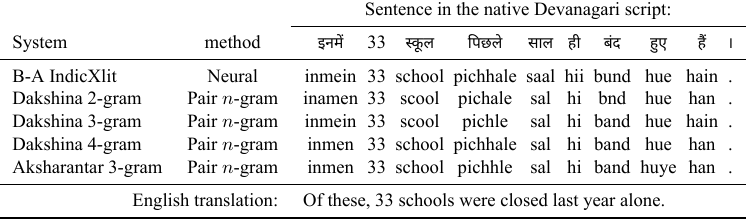}\vspace*{-0.05in}
    \caption{\footnotesize Romanizations sampled from different transliteration models.
    }\vspace*{-0.15in}
    \label{tab:fst_sample_romanizations}
\end{table*}

\subsection{Development set}
Note that, while the synthetic dev set provided by the \BhashaAbhijnaanam benchmark can be useful for tuning some system parameters, it has a critical mismatch with the test set: the test set consists of human romanized text, with varied romanizations; but the synthetic dev set consists of machine generated romanizations without such variation. In particular, the IndicXlit system romanized each native script word the same way whenever it was encountered.

In addition to evaluating on real (human) romanized text, \citet{madhani-etal-2023-bhasa} also evaluated their system's LID performance when applied to collections of synthesized romanizations, to assess the importance of the mismatch between romanizations used for training and evaluation. They show that LID accuracy improves from 80.4\% to 96.0\% when evaluating on human versus synthetic romanizations. This is unsurprising, since the distribution of synthetic romanizations are (for obvious reasons) far more similar to the training/dev sets than human romanizations are. Thus, rather than use the provided synthetic dev set, we instead chose to employ a human romanized dev set for a subset of the \BhashaAbhijnaanam languages, which we describe here.

The Dakshina dataset~\cite{roark-etal-2020-processing} consists of three parts for each of the languages in the collection: single word romanization lexicons, where native script words are paired with one or more attested romanizations;\footnote{These romanization lexicons and others are used to train romanization models used for training data synthesis.} native script Wikipedia text samples; and 10k human romanized full sentences sampled from a subset of the native script Wikipedia text. These latter full sentence romanizations are split into development and test partitions of 5k sentences each, and the \BhashaAbhijnaanam test set includes the 5k sentences from the test partition.  This leaves the 5k development partition for use as a development set, covering an eleven (out of twenty) language subset of the \BhashaAbhijnaanam task.

All hyperparameters were first tuned on this Dakshina development set, restricting the training set to the eleven languages shared with Dakshina, sampled without replacement to 1,000 examples per language.
See \cref{app:dakshina_dev_motivation} for additional discussion and results comparing the synthetic dev set with this human romanized dev set.

\subsection{Romanized training set synthesis}
The \BhashaAbhijnaanam synthesized training set for the romanized LID task was created by automatically romanizing the native script training set.  In this paper, we explore other methods for training set synthesis from the same native script training set, using pair \ngram models of various Markov orders (2--4), as described in \cref{sec:translit_background}.  These models were trained on reference input/output word pairs either from the Dakshina \cite{roark-etal-2020-processing} or Aksharantar \cite{madhani-etal-2023-aksharantar} romanization lexicons. 
In fact, significant portions of the Dakshina lexicons are directly included in the Aksharantar lexicons. However, these Dakshina subsets within Aksharantar have greater typical fertility than the rest of Aksharantar, i.e., more attested romanizations per word.
Over the 11 languages for which there are Dakshina romanization lexicons, the Aksharantar romanization lexicons contain just over 18 million unique native script words, but with just 1.011 romanizations per word.  For these same languages, the Dakshina training lexicons include 265,000 unique native script words (25k in all but Sindhi, for which there are 15k), which is less than 1.5\% of Aksharantar; however, they contain on average 2.8 romanizations per word,\footnote{The Dakshina lexicons contain between 1.7 and 4.2 romanizations per word on average, depending on the language.} hence provide many more attested romanizations per word.

Trained transliteration models are used to synthesize training sets for romanized LID from the \BhashaAbhijnaanam native script LID training data. Thus the source native script data is identical across the different synthesis methods, including the synthesized romanized training set released in the \BhashaAbhijnaanam benchmark. 
\cref{tab:fst_sample_romanizations} presents romanizations for a Hindi sentence sampled from different transliteration models.

We use two different methods to romanize a word given a model. First, we simply take the best scoring romanization of the given word, the \emph{1-best} decoded romanization.
Alternatively, we generate top-$k$ romanizations and \emph{sample} over a renormalized distribution of these, conditioned on the native script string. For this study, we extract the global 8-best romanizations and their probabilities using the Viterbi algorithm. Such exact global inference is possible since the romanization models are encoded as weighted finite state transducers. In this sense, we are sampling with temperature 1 over the set of global 8-best romanizations. Note that, when sampling from the $k$-best romanizations, each pass over the training data yields another distinctly romanized corpus. Hence, we also explore producing multiple distinctly-romanized copies of the training data. Functionality for performing such sampling is built into the transliteration utilities of the Nisaba library, see the URL in \cref{footnote:url} for details.

Both Dakshina and Aksharantar romanization lexicons omit some common symbols from their training data, including things like native script digits and punctuation, which are romanized in the baseline synthesized training data.  From this data, we identified romanized tokens that fell outside of the coverage of Dakshina or Aksharantar, and included these tokens in model training to provide equivalent coverage of the training set.



When generating synthetic training data for all 20 languages in  the \BhashaAbhijnaanam LID task, we make use of Dakshina-trained transliteration models for those languages covered by Dakshina, and Aksharantar-trained transliteration models for the rest. We find that the spelling variation induced by drawing samples from transliteration models does indeed reflect the kind of natural spelling variation reported in the literature on informal South Asian romanization. Phenomena such as implicit vowel realization, variation in indicating vowel quality, and gemination are some of the most frequent variations in these samples. See \cref{sec:analysis} and \cref{app:synthetic_spelling_variation} for analysis of spelling variability in synthetic samples.

\begin{table}[t]
    \centering%
    \small%
    \begin{tabular}{l|rr|r}
        \toprule
        & \multicolumn{2}{c|}{MADLAD} \\
        Language & noisy & clean & GlotCC \\
        \cmidrule(lr){1-1}\cmidrule(lr){2-2}\cmidrule(lr){3-3}\cmidrule(lr){4-4}
        Assamese & & & 2.0K \\
        Bengali & 1.3M & 12.0K & 2.4K \\
        Gujarati & 688.4K & 5.4K & 285 \\
        Hindi & 22.6M & 1.2M & 32.4K \\
        Kannada & 766.0K & 10.1K & 416 \\
        Konkani & & & 8.8K \\
        Malayalam & 3.5M & 77.3K & 3.5K \\
        Manipuri & & & 1.0K \\
        Marathi & & & 1.7K \\
        Nepali & & & 968  \\
        Oriya & & & 494  \\
        Punjabi & & & 4.7K  \\
        Sindhi & & & 30  \\
        Tamil & 3.4M & 142.7K & 4.5K \\
        Telugu & 4.4M  & 269.1K & 16.1K \\
        Urdu & & & 84.6K \\ \bottomrule
    \end{tabular}\vspace*{-0.05in}
    \normalsize%
    \caption{\footnotesize Number of sentences for each \BhashaAbhijnaanam romanized language in GlotCC and MADLAD. Where there is no number, there are no sentences. Bodo, Kashmiri, Maithili and Sanskrit have no romanized examples in any of these datasets.}\vspace*{-0.15in}
    \label{tab:natural_counts}
\end{table}

\subsection{Harvested natural examples}\label{sec:harvested}
Another source of romanized examples for training LID systems is already harvested text -- which we have noted is relatively sparse, but is still important to assess as a source of distant supervision.
We consider two potential sources of harvested natural romanized text. The first is MADLAD-400, a filtered subset of Common Crawl that covers a wider range of languages than mC4. The release contains text for 7 of the 20 romanized languages in B-A, with both `noisy' and `clean' sets for each. The second is GlotCC, a corpus of web documents whose language has been identified with high confidence by GlotLID, a wide coverage language identification model.
\cref{tab:natural_counts} lists the number of natural example sentences in each of these corpora for each of the \BhashaAbhijnaanam evaluation set languages.

\begin{table}[t]
  \centering%
  \small%
\begin{tabular}{@{}l@{~}l@{~~}|c@{~~~}c@{~~}c@{}}
\toprule
&& \multicolumn{3}{c}{Classifier F1}\\
\multicolumn{2}{@{}l|}{Training set} & fastText & mT5-base & mT5-large\\\midrule
\multicolumn{2}{@{}l|}{\BhashaAbhijnaanam training} &  81.4 & \textbf{84.9} & \textbf{85.3} \\\midrule
Dakshina & 2-g 1-best  & 72.5 & 75.1 & 79.2\\\cmidrule{2-5}
& 3-g 1-best  & 80.6 & 80.4 & 81.7\\\cmidrule{2-5}
& 4-g 1-best  & \textbf{83.0} & 80.9 & 82.6\\\midrule
Aksharantar & 3-g 1-best  & 78.3 & 79.6 & 78.7\\
\bottomrule
\end{tabular}
\normalsize%
\vspace*{-0.05in}
  \caption{\footnotesize Dakshina development set performance as a function of training set and model class, comparing baseline \BhashaAbhijnaanam synthesized training data with other synthesis methods that romanize every instance of a word the same (1-best).}\label{tab:onebest_results}\vspace*{-0.15in}
\end{table}

\section{Results}
In this section, we step through results on the Dakshina (human romanized) development set as we change the LID training set by using different training set synthesis models and different methods for romanizing with those models.  In each trial, we present the LID F1 score for fastText, mT5-base and mT5-large classifiers trained on the version of the training set associated with each trial, always comparing against training on the baseline training data that came with the \BhashaAbhijnaanam dataset.\footnote{We also computed accuracy for all trials, but do not report it for development set results as this metric was very similar to F1 for those trials.}

The results are structured so that the relative contribution of different key methods are demonstrated on the development set.  We begin with results using 1-best romanizations (\cref{sec:1-best-results}), then move on to demonstrate improvements due to sampling (\cref{sec:sample-results}).  We follow this by examining the impact of combining distinctly synthesized datasets (\cref{sec:combining-results}), sampling multiple copies of a dataset with the same synthesis method (\cref{sec:multiples-results}), and making use of distantly supervised corpora (\cref{sec:harvested-results}).  We then present selected system performance on the test set, and finish the section with some discussion and analysis.

\subsection{Best romanizations}\label{sec:1-best-results}
\cref{tab:onebest_results} compares systems trained on the baseline training set with those synthesized using pair \ngram transliteration models of various orders, themselves trained on either the Dakshina or Aksharantar romanization lexicons. Here the romanization for each word is the highest probability (1-best) transliteration according to the model. 

We can make a couple observations from these results.  First, none of the romanization models quite reach the performance achieved with the baseline synthesized corpus, but improvements are achieved with the Dakshina-trained models as the order of the pair \ngram models increases. The Aksharantar 3-gram model resulted in synthesized training sets that did not quite reach the level of Dakshina 3-gram model, 
despite most Dakshina romanization lexicon entries being included in 
Aksharantar for all of these languages.\footnote{We only show Pair 3-gram results for Aksharantar in the interest of conciseness, since that is the order that is eventually selected for both Dakshina and Aksharantar, for reasons that will become apparent as more results are shared.} 

\begin{table}[t]
  \centering%
  \small%
\begin{tabular}{@{}l@{~}l@{~~}|c@{~~~}c@{~~}c@{}}
\toprule
&& \multicolumn{3}{c}{Classifier F1}\\
\multicolumn{2}{@{}l|}{Training set} & fastText & mT5-base & mT5-large\\\midrule
\multicolumn{2}{@{}l|}{\BhashaAbhijnaanam training} &  81.4 & 84.9 & 85.3 \\\midrule
Dakshina & 2-g sampled  & 87.0 & 85.4 & 87.6\\\cmidrule{2-5}
& 3-g sampled & 88.0 & 86.3 & 87.3\\\cmidrule{2-5}
& 4-g sampled  & \textbf{88.5} & \textbf{87.5} & \textbf{88.4}\\\midrule
Aksharantar & 3-g sampled  & 85.6 & 84.0 & 84.9\\
\bottomrule
\end{tabular}
\normalsize%
\vspace*{-0.05in}
  \caption{\footnotesize Dakshina development set performance as a function of training set and model class, comparing baseline \BhashaAbhijnaanam synthesized training data with other synthesis methods that sample romanizations from k-best output, so that words have variable romanizations in the resulting training corpus.}\label{tab:sample_results}\vspace*{-0.15in}
\end{table}

\subsection{Sampled romanizations}\label{sec:sample-results}
\cref{tab:sample_results} compares systems trained on the baseline training set with those synthesized by sampling from pair \ngram transliteration models of various orders when trained on either Dakshina or Aksharantar romanization lexicons.  

In contrast to systems reported in \cref{tab:onebest_results}, these systems achieved large improvements over the baseline, particularly fastText systems.  For these methods, the pair \ngram model order made less difference than in \cref{tab:onebest_results}. The Aksharantar trained model again yielded synthesized training data that resulted in somewhat less performant LID systems versus the Dakshina conditions, though that condition, too, improved on the baseline.  Interestingly, for training data synthesized in this way, the best fastText system outperformed the best mT5 system.

\subsection{Combining synthetic datasets}\label{sec:combining-results}
Another source of variation beyond sampling is to combine independently synthesized training sets.  To that end, \cref{tab:union_results} presents conditions within which our newly created synthetic training sets were combined with the baseline synthetic training set, thus yielding twice the amount of text per language -- each sentence repeated twice, typically romanized distinctly.  All conditions improve from the corresponding systems reported in \cref{tab:sample_results}, though lower-order pair \ngram conditions achieved larger gains leading to similar performance across conditions. In the interest of conciseness, we focus on Pair 3-gram conditions in future results.

\begin{table}[t]
  \centering%
  \small%
\begin{tabular}{@{}l@{~}l@{~~}|c@{~~~}c@{~~}c@{}}
\toprule
\multicolumn{2}{@{}l|}{Training set unioned} & \multicolumn{3}{c}{Classifier F1}\\
\multicolumn{2}{@{}l|}{with \BhashaAbhijnaanam training} & fastText & mT5-base & mT5-large\\\midrule
\multicolumn{2}{@{}l|}{None (\BhashaAbhijnaanam alone)} &  81.4 & 84.9 & 85.3 \\\midrule
Dakshina & 2-g sampled & 88.6 & \textbf{86.6} & 87.7\\\cmidrule{2-5}
& 3-g sampled & 88.7 & 86.5 & \textbf{89.0}\\\cmidrule{2-5}
& 4-g sampled & \textbf{88.8} & \textbf{86.6} & 88.3\\\midrule
Aksharantar & 3-g sampled & 87.2 & 84.8 & 85.7\\
\bottomrule
\end{tabular}
\normalsize%
\vspace*{-0.05in}
  \caption{\footnotesize Dakshina development set performance as a function of training set and model class, comparing baseline \BhashaAbhijnaanam synthesized training data with other sampled synthesis methods when they are unioned with the baseline training data.}\label{tab:union_results}
\end{table}

\subsection{Sampling multiple copies}\label{sec:multiples-results}
If combining two distinctly romanized versions yielded some improvements, then might training on several sampled romanizations yield further gains?  Note that, when sampling from the pair \ngram models, each pass over the provided training set will yield distinctly romanized training data.  \cref{tab:copies_results} compares the baseline and 3-gram results from \cref{tab:union_results} (which used one sampled training set and the  \BhashaAbhijnaanam baseline training set), with systems trained on an additional 9 distinctly sampled romanized training sets (for a total of 10 plus the baseline). This  yielded a modest improvement for the fastText classifier in the Dakshina condition, resulting in our best result (89.2\% F1) using purely synthetic training data.  None of the mT5 model conditions improved with extra copies, nor did the Aksharantar conditions. See~\cref{app:adding_synthetic_examples} for more analysis of varying the quantity of synthetic data.

\subsection{Adding harvested data}\label{sec:harvested-results}
\begin{table}[t]
  \centering%
  \small%
\begin{tabular}{@{}l@{~}r@{~~}|c@{~~~}c@{~~}c@{}}
\toprule
\multicolumn{2}{@{}l|}{Training set unioned} & \multicolumn{3}{c}{Classifier F1}\\
\multicolumn{2}{@{}l|}{with \BhashaAbhijnaanam training} & fastText & mT5-base & mT5-large\\\midrule
\multicolumn{2}{@{}l|}{None (\BhashaAbhijnaanam alone)} &  81.4 & 84.9 & 85.3 \\\midrule
Dakshina & x 1 & 88.7 & \textbf{86.5} & \textbf{89.0}\\\cmidrule{2-5}
3-g sampled & x 10 & \textbf{89.2} & 86.4 & 88.3\\\midrule
Aksharantar &  x 1 & 87.2 & 84.8 & 85.7\\\cmidrule{2-5}
  3-g sampled & x 10 & 87.4 & 84.9 & 85.9 \\
\bottomrule
\end{tabular}
\normalsize%
\vspace*{-0.05in}
  \caption{\footnotesize Dakshina development set performance as a function of training set and model class, comparing baseline \BhashaAbhijnaanam synthesized training data with other sampled synthesis methods when they are unioned with the baseline training data. Multiple distinctly sampled versions of the training corpus can be created; here we compare the use of a single version with the use of 10 distinct versions.}\label{tab:copies_results}\vspace*{-0.15in}
\end{table}

\cref {tab:harvest_results} presents the addition of harvested text (see \cref{sec:harvested}) to earlier reported systems, including the baseline and the best synthesized training data condition.  The MADLAD-400 harvested data only degrades the baseline when added to LID training, but the GlotCC data is helpful, though not as much on its own as the improved synthetic training sets already presented -- perhaps unsurprising given the sparseness of the collection.  Combining the best harvested data set (GlotCC) with the data used in the best synthesized condition yields the best observed result by 1.3\% absolute F1.
See~\cref{app:adding_natural_examples} for more analysis of varying the quantity of harvested data.



\subsection{Test set results}

\cref{tab:ba_test} presents performance on the full \BhashaAbhijnaanam test set of published fastText and pretrained classifier baselines and systems trained for this paper using three training sets: (1) the baseline synthesized training set; (2) the best exclusively synthesized training set; and (3) the best synthesized training set combined with GlotCC harvested data. The observed patterns from the development set hold for this set as well, yielding the best reported results for this task. One notable difference between the dev and test results is that the F1 scores are universally worse than accuracy, indicating that there is some class imbalance in the predictions -- unsurprising when some of the languages have an order of magnitude less data in the test set. Still, in absolute terms, the divergence between accuracy and F1 is reduced in the best systems.

\begin{table}[t]
  \centering%
  \small%
\begin{tabular}{@{}l@{~~}|c@{~~~}c@{~~}c@{}}
\toprule
Training sets unioned & \multicolumn{3}{c}{Classifier F1}\\
with \BhashaAbhijnaanam training & fastText & mT5-base & mT5-large\\\midrule
None (\BhashaAbhijnaanam alone) &  81.4 & 84.9 & 85.3 \\\midrule
Dakshina 3g sample x 10 & 89.2 & 86.4 & 88.3\\\cmidrule{2-4}
MADLAD noisy & 80.3 & 83.2 & 84.7\\\cmidrule{2-4}
MADLAD clean & 80.7 & 82.4 & 83.6\\\cmidrule{2-4}
GlotCC & 86.5 & 85.4 & 86.7\\\cmidrule{2-4}
GlotCC $\bigcup$ Dak 3-g x 10 & \textbf{90.5} & \textbf{89.2} & \textbf{89.6}\\
\bottomrule
\end{tabular}
\normalsize%
\vspace*{-0.05in}
  \caption{\footnotesize Dakshina development set performance as a function of training set and model class, comparing baseline \BhashaAbhijnaanam synthesized training data with other sampled synthesis methods and harvested data sets when they are unioned with the baseline training data (and each other).}\label{tab:harvest_results}\vspace*{-0.15in}
\end{table}

\begin{table*}[t]
\centering%
\small%
\begin{tabular}{l|ccc}
\toprule
& fastText & \multicolumn{2}{c}{Pretrained model} \\
System & Accuracy / F1 & Accuracy / F1  & model type\\
\midrule
IndicLID \cite{madhani-etal-2023-bhasa}, published baselines & 71.5 / 63.3 & 80.4 / 74.7 & BERT \\\midrule
B-A synthetic training set & 80.7 / 71.6 & 82.4 / 73.1 & mT5-large\\
B-A + Dakshina/Aksharantar hybrid sampled x 10 & 90.5 / 85.4 & 89.1 / 83.3 & mT5-large\\
B-A + Dakshina/Aksharantar hybrid sampled x 10 + GlotCC & \bf 92.2 / 88.2 & 91.8 / 87.1 & mT5-large\\
\bottomrule
\end{tabular}\vspace*{-0.05in}
\caption{\footnotesize Accuracy and macro F1 performance on the \BhashaAbhijnaanam test set. 
}\vspace*{-0.15in}
\label{tab:ba_test}
\end{table*}

\subsection{Analysis and discussion}\label{sec:analysis}
\begin{figure}[ht!]
    \centering
    \includegraphics[width=0.46\textwidth]{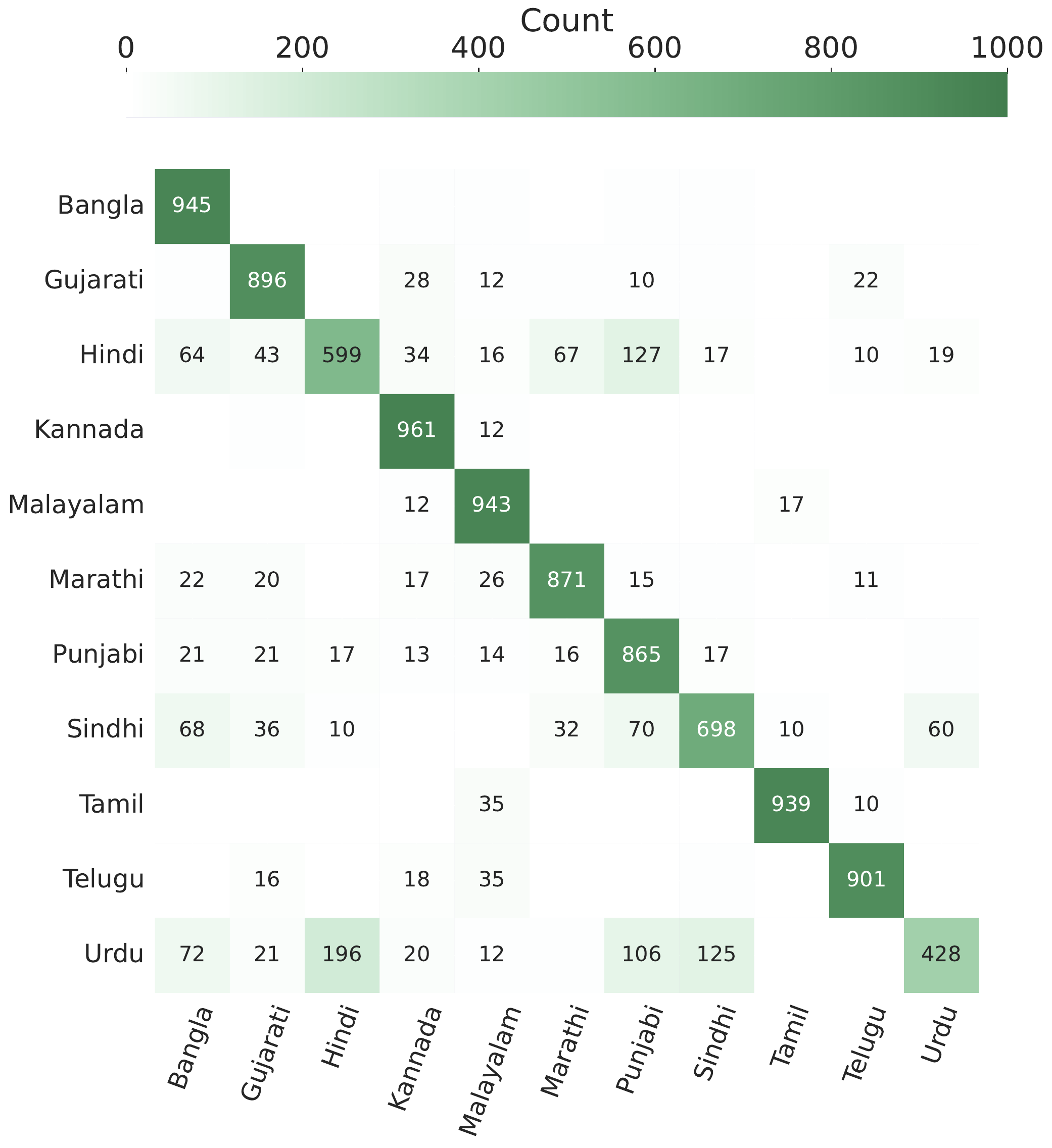}
\includegraphics[width=0.46\textwidth]{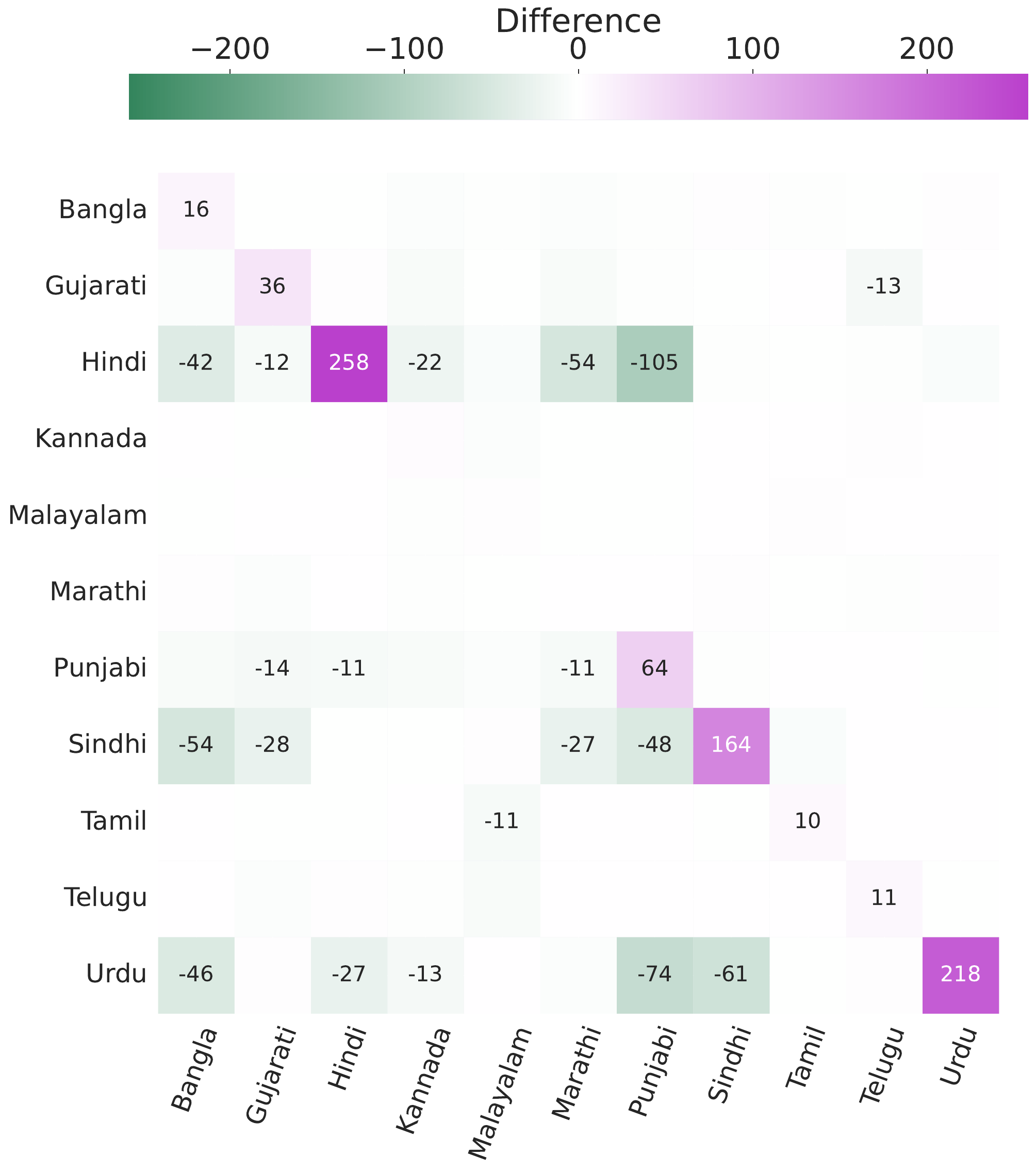}\vspace*{-0.05in}
    \caption{\footnotesize Confusion matrix for a fastText model trained on the baseline \BhashaAbhijnaanam synthetic training set (top), and improvement from training on $10\times$ additional samples from a Dakshina-trained pair \ngram transliteration model (bottom). Large, positive values on the diagonal indicate more correctly classified examples, and large, negative values in the off-diagonal entries indicate fewer confusions.}\vspace*{-0.15in}
    \label{fig:dakshina_cmax_baseline_vs_fst_10x}
\end{figure}

The best synthetic training data (Dakshina 3-gram romanizations sampled 10$\times$ plus the baseline \BhashaAbhijnaanam synthetic training set) yielded precision, recall and F1 gains for all languages in the development set, as shown in \cref{tab:dakshina_dev_per_language_fscore}. Languages which were initially poorly classified, including Hindi, Urdu, Sindhi and Punjabi, show remarkable improvements: 17.9\% absolute F1 score improvement for Hindi and 18.0\% for Urdu. 
\cref{fig:dakshina_cmax_baseline_vs_fst_10x} presents the baseline fastText system's confusion matrix on the development set, as well as a map showing where the differences fell between the baseline and the best synthetic training set. Baseline confusions are evident between (1) Hindi and Urdu; (2) Urdu and Sindhi; and (3) Punjabi and each of Hindi, Sindhi and Urdu. All of these cases improved with the updated synthetic training data. Dravidian languages (Kannada, Malayalam, Tamil, Telugu) had the best baseline performances of any languages, but still managed to achieve positive gains from the updated synthetic training data. See \cref{app:errors} for additional qualitative error analysis.

\begin{table}[t]
    \centering%
    \small%
\begin{tabular}{@{}l@{~}r@{/~}r@{/~}r@{~}|@{~~}r@{/~}r@{/~}r@{~}|@{~}r@{/}r@{/}r@{}}
\toprule
\multirow{2}{*}{Language} & \multicolumn{3}{c}{Baseline} & \multicolumn{3}{c}{Best} & \multicolumn{3}{c}{Diff} \\
 & P~ & R~ & F1 & P~ & R~ & F1 & P~ & R~ & F1 \\
\midrule
Bangla & 77.8 & 94.5 & 85.3 & 89.1 & 96.1 & 92.4 & 11.3 & 1.6 & 7.1 \\
Gujarati & 83.7 & 89.6 & 86.5 & 89.4 & 93.2 & 91.3 & 5.8 & 3.6 & 4.8 \\
Hindi & 72.2 & 59.9 & 65.5 & 81.2 & 85.7 & 83.4 & 9.0 & 25.8 & 17.9 \\
Kannada & 85.9 & 96.1 & 90.7 & 91.6 & 96.7 & 94.1 & 5.7 & 0.6 & 3.4 \\
Malayalam & 84.3 & 94.3 & 89.0 & 87.5 & 94.6 & 90.9 & 3.2 & 0.3 & 1.9 \\
Marathi & 85.5 & 87.1 & 86.3 & 96.1 & 87.3 & 91.5 & 10.7 & 0.2 & 5.2 \\
Punjabi & 71.1 & 86.5 & 78.0 & 88.6 & 92.9 & 90.7 & 17.6 & 6.4 & 12.7 \\
Sindhi & 77.6 & 69.8 & 73.5 & 86.0 & 86.2 & 86.1 & 8.5 & 16.4 & 12.6 \\
Tamil & 93.7 & 93.9 & 93.8 & 94.4 & 94.9 & 94.7 & 0.7 & 1.0 & 0.9 \\
Telugu & 91.3 & 90.1 & 90.7 & 92.7 & 91.2 & 91.9 & 1.4 & 1.1 & 1.2 \\
Urdu & 82.1 & 42.8 & 56.3 & 87.4 & 64.6 & 74.3 & 5.3 & 21.8 & 18.0 \\
\bottomrule
\end{tabular}
\normalsize%
\vspace*{-0.05in}
    \caption{\footnotesize Dakshina development set per-language precision, recall and F1 when training on (1) the baseline synthetic training set and (2) the best synthetic training set (adding Dakshina 3-gram sampled 10$\times$), as well as the absolute difference (Best minus Baseline). All differences are positive.}
    \label{tab:dakshina_dev_per_language_fscore}\vspace*{-0.15in}
\end{table}


Our hypothesis has been that sampling romanizations mimics natural spelling variation in romanized text, and for that reason using such variations in synthesized training, versus using the same (albeit likely) romanization for each instance of a word, results in better identification of the languages when written in the Latin script.  Do the samples actually mimic natural spelling variation? In an effort to answer this question, we extracted frequent alternations resulting from the sampling and coded them according to their correspondence to known alternations cited in the literature. Briefly, the most frequent alternation, accounting for nearly 50\% of the instances, corresponded to well-known variations in indication of vowel length or quality---for example, doubling of vowels to indicate length.  Another 25\% of alternations involved presence/absence of the implicit vowel, and a further 10\% the inclusion/omission of `h' to indicate aspiration or other consonant property. See \cref{app:synthetic_spelling_variation} for full details of counting and coding, along with many examples. Overall, this analysis suggests that the romanized samples that we synthesize in this work do indeed mimic spelling variation observed in natural romanizations.
\section{Conclusion and Future Work}

Through careful experimental analysis on the development set, in this paper we have demonstrated the importance of training text synthesis in improving informally romanized LID. In particular, we show that drawing samples from relatively simple romanization models yields romanizations that capture the lack of Latin orthography and spelling variability in these languages. Independently harvested text was also shown to yield further improvements, though such datasets are sparse for the languages investigated in this paper, so identifying and including more such text constitutes a major future direction. The best system -- which is the best reported result for this task by a large margin -- is a lightweight linear model, which might further benefit from feature set analysis and augmentation. Additionally, improving LID for informally romanized text outside of South Asia would be of interest, e.g., for Arabic or languages natively written using the Ge`ez script such as Amharic and Tigrinya.

\section{Limitations}
This work examines a closed-class classification task with a fixed number of labels, which would need to be modified to be applicable to a broader set of languages in real use scenarios. Additionally, the experiments examine performance on a relatively small number of languages (20) from just three language families (Indo-Aryan, Dravidian and Tibeto-Burman), and hence do not capture the diversity of languages -- including, e.g., Semitic languages -- for which informal romanization is also common.

The work focuses on dedicated LID systems with a general goal of low computational cost and latency, and does not examine the performance of commercial large language models such as ChatGPT or Gemini.


In this study we investigated LID systems in a scenario where the Latin script is used as an informal common script across various South Asian languages. 
It has not been established whether approaches that were demonstrated to be effective in this work would yield similar system improvements in scenarios where different scripts (e.g., Perso-Arabic or Devanagari) were being informally used instead of the Latin script.

\section{Ethics statement}
The goal of this work is to provide methods that advance the field's collective ability to create balanced and inclusive data sets, i.e., that include representative data from typically under-represented languages as well as from common yet chronically under-represented non-standard use scenarios, in addition to well-represented languages and conditions.  Such non-standard use scenarios may include writing in informal registers and/or with non-standard scripts or spellings, which are important forms of written communication worldwide.

\section*{Acknowledgements}
The authors would like to thank Cibu Johny and Raiomond Doctor for their help with this paper.

\bibliography{anthology,main}

\begin{thebibliography}{56}
\providecommand{\natexlab}[1]{#1}

\bibitem[{Abadji et~al.(2022)Abadji, Ortiz~Suarez, Romary, and
  Sagot}]{abadji-etal-2022-towards}
Julien Abadji, Pedro Ortiz~Suarez, Laurent Romary, and Beno{\^i}t Sagot. 2022.
\newblock \href {https://aclanthology.org/2022.lrec-1.463/} {Towards a cleaner
  document-oriented multilingual crawled corpus}.
\newblock In \emph{Proceedings of the Thirteenth Language Resources and
  Evaluation Conference}, pages 4344--4355, Marseille, France. European
  Language Resources Association.

\bibitem[{Adebara et~al.(2022)Adebara, Elmadany, Abdul-Mageed, and
  Inciarte}]{adebara-etal-2022-afrolid}
Ife Adebara, AbdelRahim Elmadany, Muhammad Abdul-Mageed, and Alcides Inciarte.
  2022.
\newblock \href {https://doi.org/10.18653/v1/2022.emnlp-main.128} {{A}fro{LID}:
  A neural language identification tool for {A}frican languages}.
\newblock In \emph{Proceedings of the 2022 Conference on Empirical Methods in
  Natural Language Processing}, pages 1958--1981, Abu Dhabi, United Arab
  Emirates. Association for Computational Linguistics.

\bibitem[{Adouane et~al.(2016)Adouane, Semmar, and
  Johansson}]{adouane-etal-2016-romanized}
Wafia Adouane, Nasredine Semmar, and Richard Johansson. 2016.
\newblock \href {https://aclanthology.org/W16-4807/} {{R}omanized {B}erber and
  {R}omanized {A}rabic automatic language identification using machine
  learning}.
\newblock In \emph{Proceedings of the Third Workshop on {NLP} for Similar
  Languages, Varieties and Dialects ({V}ar{D}ial3)}, pages 53--61, Osaka,
  Japan. The COLING 2016 Organizing Committee.

\bibitem[{Ahmed et~al.(2011)Ahmed, Bali, Choudhury, and
  VB}]{ahmed-etal-2011-challenges}
Umair~Z Ahmed, Kalika Bali, Monojit Choudhury, and Sowmya VB. 2011.
\newblock \href {https://aclanthology.org/W11-3501/} {Challenges in designing
  input method editors for {I}ndian languages: The role of word-origin and
  context}.
\newblock In \emph{Proceedings of the Workshop on Advances in Text Input
  Methods ({WTIM} 2011)}, pages 1--9, Chiang Mai, Thailand. Asian Federation of
  Natural Language Processing.

\bibitem[{Aji et~al.(2022)Aji, Winata, Koto, Cahyawijaya, Romadhony, Mahendra,
  Kurniawan, Moeljadi, Prasojo, Baldwin, Lau, and Ruder}]{aji-etal-2022-one}
Alham~Fikri Aji, Genta~Indra Winata, Fajri Koto, Samuel Cahyawijaya, Ade
  Romadhony, Rahmad Mahendra, Kemal Kurniawan, David Moeljadi, Radityo~Eko
  Prasojo, Timothy Baldwin, Jey~Han Lau, and Sebastian Ruder. 2022.
\newblock \href {https://doi.org/10.18653/v1/2022.acl-long.500} {One country,
  700+ languages: {NLP} challenges for underrepresented languages and dialects
  in {I}ndonesia}.
\newblock In \emph{Proceedings of the 60th Annual Meeting of the Association
  for Computational Linguistics (Volume 1: Long Papers)}, pages 7226--7249,
  Dublin, Ireland. Association for Computational Linguistics.

\bibitem[{Annamalai and Steever(2015)}]{annamalai:2015}
Elay Annamalai and Sanford~B. Steever. 2015.
\newblock \href {https://doi.org/10.4324/9781315722580-4} {Modern {T}amil}.
\newblock In \emph{The Dravidian Languages}, 2nd edition, Routledge Language
  Family Series, chapter~3, pages 118--175. Routledge.

\bibitem[{Bahri(2022)}]{bahri:2022}
Soubeika Bahri. 2022.
\newblock \href {https://doi.org/10.1007/978-3-031-10433-6_3} {{E}ttounsi and
  {T}amazight writing on {F}acebook: Oral vernaculars or new literacies}.
\newblock In Cecelia Cutler, May Ahmar, and Soubeika Bahri, editors,
  \emph{Digital Orality: Vernacular Writing in Online Spaces}, pages 65--93.
  Springer International Publishing, Cham, Germany.

\bibitem[{Baimukan et~al.(2022)Baimukan, Bouamor, and
  Habash}]{baimukan-etal-2022-hierarchical}
Nurpeiis Baimukan, Houda Bouamor, and Nizar Habash. 2022.
\newblock \href {https://aclanthology.org/2022.lrec-1.489/} {Hierarchical
  aggregation of dialectal data for {A}rabic dialect identification}.
\newblock In \emph{Proceedings of the Thirteenth Language Resources and
  Evaluation Conference}, pages 4586--4596, Marseille, France. European
  Language Resources Association.

\bibitem[{Baruah et~al.(2024)Baruah, Singh, and
  Sarmah}]{baruah2024transliteration}
Hemanta Baruah, Sanasam~Ranbir Singh, and Priyankoo Sarmah. 2024.
\newblock \href {https://doi.org/10.1145/3639565} {Transliteration
  characteristics in romanized {A}ssamese language social media text and
  machine transliteration}.
\newblock \emph{ACM Transactions on Asian and Low-Resource Language Information
  Processing}, 23(2):1--36.

\bibitem[{Beesley(1988)}]{beesley1988language}
Kenneth~R Beesley. 1988.
\newblock Language identifier: A computer program for automatic
  natural-language identification of on-line text.
\newblock In \emph{Proceedings of the 29th annual conference of the American
  Translators Association}, volume~47, page~54.

\bibitem[{Bhatia(2010)}]{bhatia:2010}
Tej~Krishan Bhatia. 2010.
\newblock \href {https://www.routledge.com/Punjabi/Bhatia/p/book/9780415589932}
  {\emph{Punjabi: A cognitive-descriptive grammar}}.
\newblock Descriptive Grammars. Routledge.

\bibitem[{Bisani and Ney(2008)}]{bisani2008joint}
Maximilian Bisani and Hermann Ney. 2008.
\newblock \href {https://doi.org/10.1016/j.specom.2008.01.002} {Joint-sequence
  models for grapheme-to-phoneme conversion}.
\newblock \emph{Speech Communication}, 50(5):434--451.

\bibitem[{Brandt(2020)}]{brandt:2020}
Carmen Brandt. 2020.
\newblock \href {https://doi.org/10.11588/xabooks.642} {From a symbol of
  colonial conquest to the {\emph{scripta} \emph{franca}}: The {R}oman script
  for {S}outh {A}sian languages}.
\newblock In Carmen Brandt and Hans Harder, editors, \emph{Wege durchs
  {L}abyrinth: {F}estschrift zu {E}hren von {R}ahul {P}eter {D}as}, pages
  1--36. CrossAsia-eBooks, Heidelberg, Germany.

\bibitem[{Brown(2014)}]{brown-2014-non}
Ralf Brown. 2014.
\newblock \href {https://doi.org/10.3115/v1/D14-1069} {Non-linear mapping for
  improved identification of 1300+ languages}.
\newblock In \emph{Proceedings of the 2014 Conference on Empirical Methods in
  Natural Language Processing ({EMNLP})}, pages 627--632, Doha, Qatar.
  Association for Computational Linguistics.

\bibitem[{Burchell et~al.(2023)Burchell, Birch, Bogoychev, and
  Heafield}]{burchell-etal-2023-open}
Laurie Burchell, Alexandra Birch, Nikolay Bogoychev, and Kenneth Heafield.
  2023.
\newblock \href {https://doi.org/10.18653/v1/2023.acl-short.75} {An open
  dataset and model for language identification}.
\newblock In \emph{Proceedings of the 61st Annual Meeting of the Association
  for Computational Linguistics (Volume 2: Short Papers)}, pages 865--879,
  Toronto, Canada. Association for Computational Linguistics.

\bibitem[{Caswell et~al.(2020)Caswell, Breiner, van Esch, and
  Bapna}]{caswell-etal-2020-language}
Isaac Caswell, Theresa Breiner, Daan van Esch, and Ankur Bapna. 2020.
\newblock \href {https://doi.org/10.18653/v1/2020.coling-main.579} {Language
  {ID} in the wild: Unexpected challenges on the path to a thousand-language
  web text corpus}.
\newblock In \emph{Proceedings of the 28th International Conference on
  Computational Linguistics}, pages 6588--6608, Barcelona, Spain (Online).
  International Committee on Computational Linguistics.

\bibitem[{Cavnar and Trenkle(1994)}]{cavnar1994n}
William~B Cavnar and John~M Trenkle. 1994.
\newblock N-gram-based text categorization.
\newblock In \emph{Proceedings of SDAIR-94, 3rd Annual Symposium on Document
  Analysis and Information Retrieval}, volume 161175, page~14, Las Vegas, NV.

\bibitem[{Chen et~al.(2024)Chen, Adebara, Doan, Liao, and
  Abdul-Mageed}]{chen-etal-2024-fumbling}
Wei-Rui Chen, Ife Adebara, Khai Doan, Qisheng Liao, and Muhammad Abdul-Mageed.
  2024.
\newblock \href {https://doi.org/10.18653/v1/2024.findings-naacl.274} {Fumbling
  in {B}abel: An investigation into {C}hat{GPT}`s language identification
  ability}.
\newblock In \emph{Findings of the Association for Computational Linguistics:
  NAACL 2024}, pages 4387--4413, Mexico City, Mexico. Association for
  Computational Linguistics.

\bibitem[{Choksi(2020)}]{choksi:2020}
Nishaant Choksi. 2020.
\newblock \href {https://doi.org/10.1086/706549} {From transcript to
  ``trans-script'': Romanized {S}antali across semiotic media}.
\newblock \emph{Signs and Society}, 8(1):62--92.

\bibitem[{Church(1986)}]{church1986stress}
Kenneth Church. 1986.
\newblock \href {https://doi.org/10.1109/ICASSP.1986.1169265} {Stress
  assignment in letter to sound rules for speech synthesis}.
\newblock In \emph{ICASSP'86. IEEE International Conference on Acoustics,
  Speech, and Signal Processing}, volume~11, pages 2423--2426, Tokyo, Japan.
  IEEE.

\bibitem[{Ciotti(2017)}]{ciotti:2017}
Giovanni Ciotti. 2017.
\newblock \href {https://doi.org/10.1051/hel/2017390205} {On (the)
  {\emph{sandhi}} between the {T}amil and {S}anskrit grammatical traditions}.
\newblock \emph{Histoire Epist{\'e}mologie Langage}, 39(2):89--102.

\bibitem[{{\c{C}}{\"o}ltekin et~al.(2018){\c{C}}{\"o}ltekin, Rama, and
  Blaschke}]{coltekin-etal-2018-tubingen}
{\c{C}}a{\u{g}}r{\i} {\c{C}}{\"o}ltekin, Taraka Rama, and Verena Blaschke.
  2018.
\newblock \href {https://aclanthology.org/W18-3906/} {{T}{\"u}bingen-{O}slo
  team at the {V}ar{D}ial 2018 evaluation campaign: An analysis of n-gram
  features in language variety identification}.
\newblock In \emph{Proceedings of the Fifth Workshop on {NLP} for Similar
  Languages, Varieties and Dialects ({V}ar{D}ial 2018)}, pages 55--65, Santa
  Fe, New Mexico, USA. Association for Computational Linguistics.

\bibitem[{Conneau et~al.(2020)Conneau, Khandelwal, Goyal, Chaudhary, Wenzek,
  Guzm{\'a}n, Grave, Ott, Zettlemoyer, and
  Stoyanov}]{conneau-etal-2020-unsupervised}
Alexis Conneau, Kartikay Khandelwal, Naman Goyal, Vishrav Chaudhary, Guillaume
  Wenzek, Francisco Guzm{\'a}n, Edouard Grave, Myle Ott, Luke Zettlemoyer, and
  Veselin Stoyanov. 2020.
\newblock \href {https://doi.org/10.18653/v1/2020.acl-main.747} {Unsupervised
  cross-lingual representation learning at scale}.
\newblock In \emph{Proceedings of the 58th Annual Meeting of the Association
  for Computational Linguistics}, pages 8440--8451, Online. Association for
  Computational Linguistics.

\bibitem[{Cortes and Vapnik(1995)}]{cortes:1995}
Corinna Cortes and Vladimir Vapnik. 1995.
\newblock \href {https://doi.org/10.1007/BF00994018} {Support-vector networks}.
\newblock \emph{Machine Learning}, 20:273--297.

\bibitem[{Demirsahin et~al.(2022)Demirsahin, Johny, Gutkin, and
  Roark}]{demirsahin-etal-2022-criteria}
Isin Demirsahin, Cibu Johny, Alexander Gutkin, and Brian Roark. 2022.
\newblock \href {https://aclanthology.org/2022.lrec-1.718/} {Criteria for
  useful automatic {R}omanization in {S}outh {A}sian languages}.
\newblock In \emph{Proceedings of the Thirteenth Language Resources and
  Evaluation Conference}, pages 6662--6673, Marseille, France. European
  Language Resources Association.

\bibitem[{Devlin et~al.(2019)Devlin, Chang, Lee, and
  Toutanova}]{devlin-etal-2019-bert}
Jacob Devlin, Ming-Wei Chang, Kenton Lee, and Kristina Toutanova. 2019.
\newblock \href {https://doi.org/10.18653/v1/N19-1423} {{BERT}: Pre-training of
  deep bidirectional transformers for language understanding}.
\newblock In \emph{Proceedings of the 2019 Conference of the North {A}merican
  Chapter of the Association for Computational Linguistics: Human Language
  Technologies, Volume 1 (Long and Short Papers)}, pages 4171--4186,
  Minneapolis, Minnesota. Association for Computational Linguistics.

\bibitem[{Dey et~al.(2024)Dey, Thakur, Kandwal, Kumar, Dasgupta, and
  Pratim~Roy}]{dey:2024}
Sayantan Dey, Shivam Thakur, Akhilesh Kandwal, Rohit Kumar, Sharmistha
  Dasgupta, and Partha Pratim~Roy. 2024.
\newblock \href {https://doi.org/10.1109/ACCESS.2024.3396290} {{BharatBhasaNet}
  - a unified framework to identify {I}ndian code mix languages}.
\newblock \emph{IEEE Access}, 12:68893--68904.

\bibitem[{Hellsten et~al.(2017)Hellsten, Roark, Goyal, Allauzen, Beaufays,
  Ouyang, Riley, and Rybach}]{hellsten-etal-2017-transliterated}
Lars Hellsten, Brian Roark, Prasoon Goyal, Cyril Allauzen, Fran{\c{c}}oise
  Beaufays, Tom Ouyang, Michael Riley, and David Rybach. 2017.
\newblock \href {https://doi.org/10.18653/v1/W17-4002} {Transliterated mobile
  keyboard input via weighted finite-state transducers}.
\newblock In \emph{Proceedings of the 13th International Conference on Finite
  State Methods and Natural Language Processing ({FSMNLP} 2017)}, pages 10--19,
  Ume{\r{a}}, Sweden. Association for Computational Linguistics.

\bibitem[{Irvine et~al.(2012)Irvine, Weese, and
  Callison-Burch}]{irvine-etal-2012-processing}
Ann Irvine, Jonathan Weese, and Chris Callison-Burch. 2012.
\newblock \href {https://aclanthology.org/W12-2109/} {Processing informal,
  {R}omanized {P}akistani text messages}.
\newblock In \emph{Proceedings of the Second Workshop on Language in Social
  Media}, pages 75--78, Montr{\'e}al, Canada. Association for Computational
  Linguistics.

\bibitem[{Jauhiainen et~al.(2019)Jauhiainen, Lui, Zampieri, Baldwin, and
  Lind{\'e}n}]{jauhiainen2019automatic}
Tommi Jauhiainen, Marco Lui, Marcos Zampieri, Timothy Baldwin, and Krister
  Lind{\'e}n. 2019.
\newblock \href {https://www.jair.org/index.php/jair/article/view/11675/26513}
  {Automatic language identification in texts: A survey}.
\newblock \emph{Journal of Artificial Intelligence Research}, 65:675--782.

\bibitem[{Joulin et~al.(2016)Joulin, Grave, Bojanowski, Douze, J{\'e}gou, and
  Mikolov}]{joulin2016fasttext}
Armand Joulin, Edouard Grave, Piotr Bojanowski, Matthijs Douze, H{\'e}rve
  J{\'e}gou, and Tomas Mikolov. 2016.
\newblock \href {https://doi.org/10.48550/arXiv.1612.03651} {{F}ast{T}ext.zip:
  Compressing text classification models}.
\newblock \emph{arXiv preprint arXiv:1612.03651}.

\bibitem[{Joulin et~al.(2017)Joulin, Grave, Bojanowski, and
  Mikolov}]{joulin-etal-2017-bag}
Armand Joulin, Edouard Grave, Piotr Bojanowski, and Tomas Mikolov. 2017.
\newblock \href {https://aclanthology.org/E17-2068/} {Bag of tricks for
  efficient text classification}.
\newblock In \emph{Proceedings of the 15th Conference of the {E}uropean Chapter
  of the Association for Computational Linguistics: Volume 2, Short Papers},
  pages 427--431, Valencia, Spain. Association for Computational Linguistics.

\bibitem[{Kargaran et~al.(2023)Kargaran, Imani, Yvon, and
  Schuetze}]{kargaran-etal-2023-glotlid}
Amir~Hossein Kargaran, Ayyoob Imani, Fran{\c{c}}ois Yvon, and Hinrich Schuetze.
  2023.
\newblock \href {https://doi.org/10.18653/v1/2023.findings-emnlp.410}
  {{G}lot{LID}: Language identification for low-resource languages}.
\newblock In \emph{Findings of the Association for Computational Linguistics:
  EMNLP 2023}, pages 6155--6218, Singapore. Association for Computational
  Linguistics.

\bibitem[{Kargaran et~al.(2024)Kargaran, Yvon, and
  Sch{\"u}tze}]{kargaran2024glotcc}
Amir~Hossein Kargaran, Fran{\c{c}}ois Yvon, and Hinrich Sch{\"u}tze. 2024.
\newblock \href {https://openreview.net/forum?id=aJ1yse8GEr} {Glot{CC}: An open
  broad-coverage {CommonCrawl} corpus and pipeline for minority languages}.
\newblock In \emph{Proceedings of the 38th Conference on Neural Information
  Processing Systems (NeurIPS), Datasets and Benchmarks Track}, Vancouver,
  Canada.

\bibitem[{Kirov et~al.(2024)Kirov, Johny, Katanova, Gutkin, and
  Roark}]{kirov-etal-2024-context}
Christo Kirov, Cibu Johny, Anna Katanova, Alexander Gutkin, and Brian Roark.
  2024.
\newblock \href {https://doi.org/10.1162/coli_a_00510} {Context-aware
  transliteration of {R}omanized {S}outh {A}sian languages}.
\newblock \emph{Computational Linguistics}, 50(2):475--534.

\bibitem[{Kreutzer et~al.(2022)Kreutzer, Caswell, Wang, Wahab, van Esch,
  Ulzii-Orshikh, Tapo, Subramani, Sokolov, Sikasote, Setyawan, Sarin, Samb,
  Sagot, Rivera, Rios, Papadimitriou, Osei, Suarez, Orife, Ogueji, Rubungo,
  Nguyen, M{\"u}ller, M{\"u}ller, Muhammad, Muhammad, Mnyakeni, Mirzakhalov,
  Matangira, Leong, Lawson, Kudugunta, Jernite, Jenny, Firat, Dossou, Dlamini,
  de~Silva, {\c{C}}abuk~Ball{\i}, Biderman, Battisti, Baruwa, Bapna, Baljekar,
  Azime, Awokoya, Ataman, Ahia, Ahia, Agrawal, and
  Adeyemi}]{kreutzer-etal-2022-quality}
Julia Kreutzer, Isaac Caswell, Lisa Wang, Ahsan Wahab, Daan van Esch,
  Nasanbayar Ulzii-Orshikh, Allahsera Tapo, Nishant Subramani, Artem Sokolov,
  Claytone Sikasote, Monang Setyawan, Supheakmungkol Sarin, Sokhar Samb,
  Beno{\^i}t Sagot, Clara Rivera, Annette Rios, Isabel Papadimitriou, Salomey
  Osei, Pedro~Ortiz Suarez, Iroro Orife, Kelechi Ogueji, Andre~Niyongabo
  Rubungo, Toan~Q. Nguyen, Mathias M{\"u}ller, Andr{\'e} M{\"u}ller,
  Shamsuddeen~Hassan Muhammad, Nanda Muhammad, Ayanda Mnyakeni, Jamshidbek
  Mirzakhalov, Tapiwanashe Matangira, Colin Leong, Nze Lawson, Sneha Kudugunta,
  Yacine Jernite, Mathias Jenny, Orhan Firat, Bonaventure F.~P. Dossou, Sakhile
  Dlamini, Nisansa de~Silva, Sakine {\c{C}}abuk~Ball{\i}, Stella Biderman,
  Alessia Battisti, Ahmed Baruwa, Ankur Bapna, Pallavi Baljekar, Israel~Abebe
  Azime, Ayodele Awokoya, Duygu Ataman, Orevaoghene Ahia, Oghenefego Ahia,
  Sweta Agrawal, and Mofetoluwa Adeyemi. 2022.
\newblock \href {https://doi.org/10.1162/tacl_a_00447} {Quality at a glance: An
  audit of web-crawled multilingual datasets}.
\newblock \emph{Transactions of the Association for Computational Linguistics},
  10:50--72.

\bibitem[{Kudo and Richardson(2018)}]{kudo-richardson-2018-sentencepiece}
Taku Kudo and John Richardson. 2018.
\newblock \href {https://doi.org/10.18653/v1/D18-2012} {{S}entence{P}iece: A
  simple and language independent subword tokenizer and detokenizer for neural
  text processing}.
\newblock In \emph{Proceedings of the 2018 Conference on Empirical Methods in
  Natural Language Processing: System Demonstrations}, pages 66--71, Brussels,
  Belgium. Association for Computational Linguistics.

\bibitem[{Kudugunta et~al.(2023)Kudugunta, Caswell, Zhang, Garcia, Xin,
  Kusupati, Stella, Bapna, and Firat}]{kudugunta:2023}
Sneha Kudugunta, Isaac Caswell, Biao Zhang, Xavier Garcia, Derrick Xin, Aditya
  Kusupati, Romi Stella, Ankur Bapna, and Orhan Firat. 2023.
\newblock \href
  {https://proceedings.neurips.cc/paper_files/paper/2023/file/d49042a5d49818711c401d34172f9900-Paper-Datasets_and_Benchmarks.pdf}
  {{MADLAD}-400: A multilingual and document-level large audited dataset}.
\newblock \emph{Advances in Neural Information Processing Systems (NeurIPS)},
  36:67284--67296.

\bibitem[{Lui and Baldwin(2012)}]{lui-baldwin-2012-langid}
Marco Lui and Timothy Baldwin. 2012.
\newblock \href {https://aclanthology.org/P12-3005/} {langid.py: An
  off-the-shelf language identification tool}.
\newblock In \emph{Proceedings of the {ACL} 2012 System Demonstrations}, pages
  25--30, Jeju Island, Korea. Association for Computational Linguistics.

\bibitem[{Madhani et~al.(2023{\natexlab{a}})Madhani, Khapra, and
  Kunchukuttan}]{madhani-etal-2023-bhasa}
Yash Madhani, Mitesh~M. Khapra, and Anoop Kunchukuttan. 2023{\natexlab{a}}.
\newblock \href {https://doi.org/10.18653/v1/2023.acl-short.71}
  {Bhasa-{A}bhijnaanam: Native-script and romanized language identification for
  22 {I}ndic languages}.
\newblock In \emph{Proceedings of the 61st Annual Meeting of the Association
  for Computational Linguistics (Volume 2: Short Papers)}, pages 816--826,
  Toronto, Canada. Association for Computational Linguistics.

\bibitem[{Madhani et~al.(2023{\natexlab{b}})Madhani, Parthan, Bedekar, Nc,
  Khapra, Kunchukuttan, Kumar, and Khapra}]{madhani-etal-2023-aksharantar}
Yash Madhani, Sushane Parthan, Priyanka Bedekar, Gokul Nc, Ruchi Khapra, Anoop
  Kunchukuttan, Pratyush Kumar, and Mitesh Khapra. 2023{\natexlab{b}}.
\newblock \href {https://doi.org/10.18653/v1/2023.findings-emnlp.4}
  {Aksharantar: Open {I}ndic-language transliteration datasets and models for
  the next billion users}.
\newblock In \emph{Findings of the Association for Computational Linguistics:
  EMNLP 2023}, pages 40--57, Singapore. Association for Computational
  Linguistics.

\bibitem[{Manukonda and Kodali(2025)}]{manukonda-kodali-2025-bytesizedllm}
Durga~Prasad Manukonda and Rohith~Gowtham Kodali. 2025.
\newblock \href {https://aclanthology.org/2025.chipsal-1.26/}
  {byte{S}ized{LLM}@{NLU} of {D}evanagari script languages 2025: Language
  identification using customized attention {B}i{LSTM} and {XLM}-{R}o{BERT}a
  base embeddings}.
\newblock In \emph{Proceedings of the First Workshop on Challenges in
  Processing South Asian Languages (CHiPSAL 2025)}, pages 248--252, Abu Dhabi,
  UAE. International Committee on Computational Linguistics.

\bibitem[{Mohan et~al.(2023)Mohan, Kumar, Elakkiya, Venkatakrishnan, Harrieni,
  Harshitha, Harini, and Reddy}]{mohan:2023}
G.~Bharathi Mohan, R.~Prasanna Kumar, R.~Elakkiya, R.~Venkatakrishnan, Shankar
  Harrieni, Y.~Sree Harshitha, K.~Harini, and M.~Nikhil Reddy. 2023.
\newblock \href {https://doi.org/10.1109/ICSCAN58655.2023.10394757}
  {Transformer-based models for language identification: A comparative study}.
\newblock In \emph{2023 International Conference on System, Computation,
  Automation and Networking (ICSCAN)}, pages 1--6, Puducherry, India. IEEE.

\bibitem[{Mustonen(1965)}]{mustonen:1965}
Seppo Mustonen. 1965.
\newblock Multiple discriminant analysis in linguistic problems.
\newblock \emph{Statistical Methods in Linguistics}, 4:37--44.

\bibitem[{Nielsen et~al.(2023)Nielsen, Kirov, and
  Roark}]{nielsen-etal-2023-distinguishing}
Elizabeth Nielsen, Christo Kirov, and Brian Roark. 2023.
\newblock \href {https://doi.org/10.18653/v1/2023.cawl-1.5} {Distinguishing
  {R}omanized {H}indi from {R}omanized {U}rdu}.
\newblock In \emph{Proceedings of the Workshop on Computation and Written
  Language (CAWL 2023)}, pages 33--42, Toronto, Canada. Association for
  Computational Linguistics.

\bibitem[{Pavan et~al.(2010)Pavan, Tandon, and Varma}]{pavan:2010}
Kosuru Pavan, Niket Tandon, and Vasudeva Varma. 2010.
\newblock \href
  {https://citeseerx.ist.psu.edu/document?repid=rep1&type=pdf&doi=f96237bda9d00ce1dc81897a9b29d05791cf99ff}
  {Addressing challenges in automatic language identification of romanized
  text}.
\newblock In \emph{8th International Conference on Natural Language Processing
  (ICON 2010)}, Kharagpur, India.

\bibitem[{Perera et~al.(2025)Perera, Jayakodi, Sumanathilaka, and
  Anuradha}]{perera-etal-2025-indonlp}
Sandun~Sameera Perera, Lahiru~Prabhath Jayakodi, Deshan~Koshala Sumanathilaka,
  and Isuri Anuradha. 2025.
\newblock \href {https://aclanthology.org/2025.indonlp-1.16/} {{I}ndo{NLP} 2025
  shared task: {R}omanized {S}inhala to {S}inhala reverse transliteration using
  {BERT}}.
\newblock In \emph{Proceedings of the First Workshop on Natural Language
  Processing for Indo-Aryan and Dravidian Languages}, pages 135--140, Abu
  Dhabi. Association for Computational Linguistics.

\bibitem[{Pierrehumbert and Nair(1996)}]{pierrehumbert:1996}
Janet Pierrehumbert and Rami Nair. 1996.
\newblock Implications of {H}indi prosodic structure.
\newblock \emph{Current trends in phonology: Models and methods}, 2:549--584.

\bibitem[{Post and Burling(2017)}]{post:2017}
Mark~W. Post and Robbins Burling. 2017.
\newblock The {T}ibeto-{B}urman languages of {N}ortheast {I}ndia.
\newblock In Graham Thurgood and Randy~J. LaPolla, editors, \emph{The
  Sino-Tibetan Languages}, 2nd edition, Language Family Series. Routledge,
  London.

\bibitem[{Raffel et~al.(2020)Raffel, Shazeer, Roberts, Lee, Narang, Matena,
  Zhou, Li, and Liu}]{raffel:2020}
Colin Raffel, Noam Shazeer, Adam Roberts, Katherine Lee, Sharan Narang, Michael
  Matena, Yanqi Zhou, Wei Li, and Peter~J. Liu. 2020.
\newblock \href {https://dl.acm.org/doi/10.5555/3455716.3455856} {Exploring the
  limits of transfer learning with a unified text-to-text transformer}.
\newblock \emph{Journal of Machine Learning Research}, 21(1):5485--5551.

\bibitem[{Roark et~al.(2020)Roark, Wolf-Sonkin, Kirov, Mielke, Johny,
  Demirsahin, and Hall}]{roark-etal-2020-processing}
Brian Roark, Lawrence Wolf-Sonkin, Christo Kirov, Sabrina~J. Mielke, Cibu
  Johny, Isin Demirsahin, and Keith Hall. 2020.
\newblock \href {https://aclanthology.org/2020.lrec-1.294/} {Processing {S}outh
  {A}sian languages written in the {L}atin script: the {D}akshina dataset}.
\newblock In \emph{Proceedings of the Twelfth Language Resources and Evaluation
  Conference}, pages 2413--2423, Marseille, France. European Language Resources
  Association.

\bibitem[{Schiffman(1999)}]{schiffman:1999}
Harold~F. Schiffman. 1999.
\newblock \emph{A reference grammar of spoken Tamil}.
\newblock Reference Grammars. Cambridge University Press, Cambridge, UK.

\bibitem[{Toftrup et~al.(2021)Toftrup, Asger~S{\o}rensen, Ciosici, and
  Assent}]{toftrup-etal-2021-reproduction}
Mads Toftrup, S{\o}ren Asger~S{\o}rensen, Manuel~R. Ciosici, and Ira Assent.
  2021.
\newblock \href {https://doi.org/10.18653/v1/2021.eacl-srw.6} {A reproduction
  of apple`s bi-directional {LSTM} models for language identification in short
  strings}.
\newblock In \emph{Proceedings of the 16th Conference of the European Chapter
  of the Association for Computational Linguistics: Student Research Workshop},
  pages 36--42, Online. Association for Computational Linguistics.

\bibitem[{Torwali(2020)}]{torwali:2020}
Zubair Torwali. 2020.
\newblock \href {https://www.lddjournal.org/article/id/1220/} {Countering the
  challenges of globalization faced by endangered languages of {N}orth
  {P}akistan}.
\newblock \emph{Language Documentation and Description}, 17:44--65.

\bibitem[{Wolf-Sonkin et~al.(2019)Wolf-Sonkin, Schogol, Roark, and
  Riley}]{wolf-sonkin-etal-2019-latin}
Lawrence Wolf-Sonkin, Vlad Schogol, Brian Roark, and Michael Riley. 2019.
\newblock \href {https://doi.org/10.18653/v1/W19-3114} {{L}atin script
  keyboards for {S}outh {A}sian languages with finite-state normalization}.
\newblock In \emph{Proceedings of the 14th International Conference on
  Finite-State Methods and Natural Language Processing}, pages 108--117,
  Dresden, Germany. Association for Computational Linguistics.

\bibitem[{Xue et~al.(2021)Xue, Constant, Roberts, Kale, Al-Rfou, Siddhant,
  Barua, and Raffel}]{xue-etal-2021-mt5}
Linting Xue, Noah Constant, Adam Roberts, Mihir Kale, Rami Al-Rfou, Aditya
  Siddhant, Aditya Barua, and Colin Raffel. 2021.
\newblock \href {https://doi.org/10.18653/v1/2021.naacl-main.41} {m{T}5: A
  massively multilingual pre-trained text-to-text transformer}.
\newblock In \emph{Proceedings of the 2021 Conference of the North American
  Chapter of the Association for Computational Linguistics: Human Language
  Technologies}, pages 483--498, Online. Association for Computational
  Linguistics.

\end{thebibliography}


\appendix
\section{LID Training details}
\label{app:langid_training}

\begin{table}[t]
\centering%
\small%
\npdecimalsign{.}
\nprounddigits{1}
\begin{tabular}{cccn{2}{2}n{2}{2}}
\toprule
{\multirow{2}{*}{No Punct}} & \multicolumn{2}{l}{Char \ngram order} & {\multirow{2}{*}{Accuracy}} & {\multirow{2}{*}{Macro F1}} \\
& {Min} & {Max} & & \\
\cmidrule(lr){1-1}\cmidrule(lr){2-3}\cmidrule(lr){4-4}\cmidrule(lr){5-5}
 \xmark & 1 & 3 & 81.0818 & 80.4137 \\
 \checkmark & 1 & 3 & 81.5000 & 80.9157 \\
 \checkmark & 2 & 4 & 81.2818 & 80.4575 \\
 \checkmark & 3 & 4 & 80.9727 & 80.1269 \\
 \checkmark & 1 & 5 & 81.8909 & 81.0661 \\
 \checkmark & 2 & 5 & 81.5545 & 80.7105 \\
 \checkmark & 3 & 5 & 81.7727 & 80.9466 \\
 \checkmark & 1 & 6 & 82.0636 & 81.2474 \\
 \checkmark & 2 & 6 & 81.9000 & 81.1152 \\
 \checkmark & 3 & 6 & 81.8545 & 81.0146 \\
 \checkmark & 3 & 7 & 82.2364 & 81.3829 \\
 \checkmark & 3 & 8 & 82.1091 & 81.3121 \\
 \checkmark & 3 & 9 & {\npboldmath} 82.2818 & {\npboldmath} 81.4834 \\
 \checkmark & 4 & 7 & 81.9818 & 81.2985 \\
 \checkmark & 4 & 8 & 82.1455 & {\npboldmath} 81.4850 \\
 \checkmark & 4 & 9 & 82.0182 & 81.3345 \\
\bottomrule
\normalsize%
\end{tabular}\vspace*{-0.05in}
    \caption{\footnotesize fastText Dakshina development set \% performance as a function of hyperparameters. Models are trained on the released B-A romanized training set restricted to the Dakshina languages, with hidden layer dimension 16. We selected character \ngrams in [3, 7], since we found that setting performed well, with further increase to the min/max character \ngram value yielding marginal performance gain.}\vspace*{-0.15in}
    \label{tab:fasttext_hyperparameter_sweep}
\end{table}

All neural models were finetuned with a constant learning rate of $10^{-3}$ for 50,000 iterations of batch size 64, with an input sequence length of 256 SentencePiece tokens~\cite{kudo-richardson-2018-sentencepiece}. This matches the finetuning described in \citet{xue-etal-2021-mt5}, and took 164.3 TensorCore-hours on a Cloud TPU v3\footnote{https://cloud.google.com/tpu/docs/v3} for an mT5-large model. FastText models were trained directly on the supervised training set (no unsupervised pretraining), with hidden dimension 16, and all character \ngrams in the range [3, 7]. 

\cref{tab:fasttext_hyperparameter_sweep} presents development set accuracy and F1 as these metaparameters were varied. We removed all non-alphanumeric characters as part of our preprocessing, as we found that these features did not generalize well on the development set. Initially we found that the fastText models picked up on punctuation as being indicative of Kashmiri -- possibly an artifact of the domain from which the Kashmiri examples were sourced. FastText training is cheap, on the order of minutes purely using CPU.

\section{Varying the number of synthetic training examples}
\label{app:adding_synthetic_examples}
\begin{figure}[t]
    \centering
    \includegraphics[width=.96\linewidth]{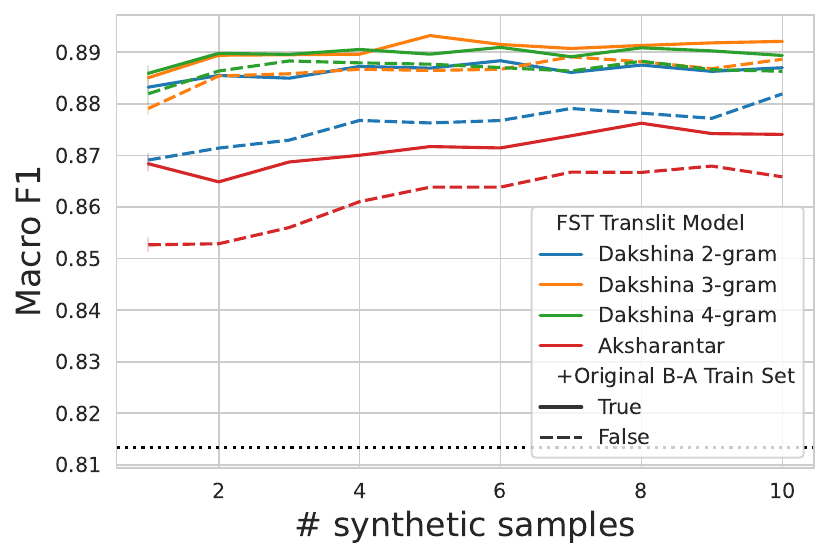}\vspace*{-0.05in}
    \caption{\footnotesize Macro F1 for various synthetic training sets as a function of number of samples to train a fastText LID model on. Solid lines indicate that the LID model was also trained on the original \BhashaAbhijnaanam data, while a dotted line indicates only training on synthesized samples from the pair \ngram model. Baseline performance is indicated by the dotted black line at the bottom.}\vspace*{-0.15in}
    \label{fig:vary_synthetic_examples}
\end{figure}

\begin{figure}[t]
    \centering
    \includegraphics[width=.96\linewidth]{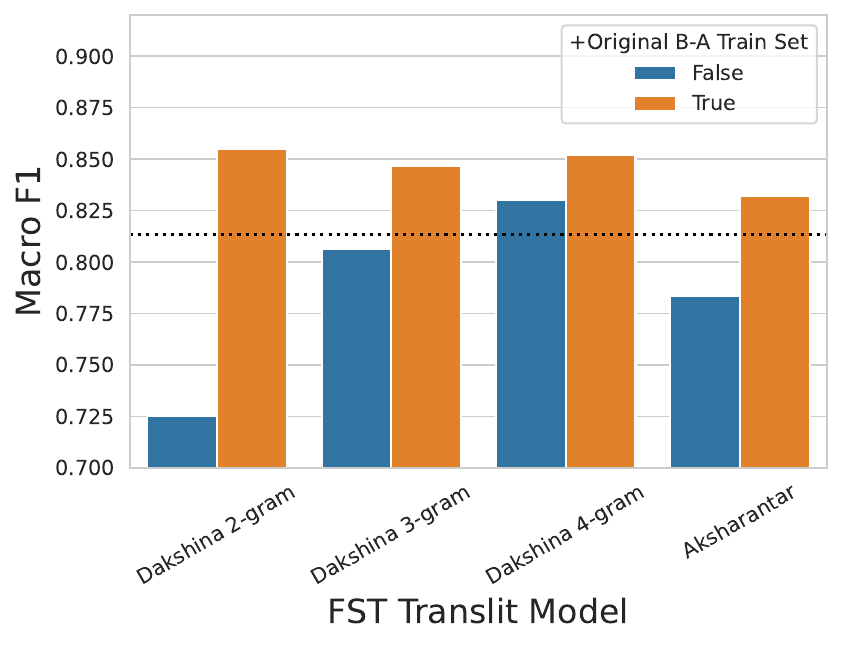}\vspace*{-0.05in}
    \caption{\footnotesize Macro F1 for various synthetic training sets, when decoding the 1-best candidate for each native word under the pair \ngram transliteration model, rather than sampling from the 8-best candidate list. The dotted black line indicates baseline performance of training on the original synthetic training set.}\vspace*{-0.15in}
    \label{fig:synthetic_1best}
\end{figure}


For all orders of pair \ngram models, synthesizing more sampled training data tended to improve development set performance, however these gains were marginal relative to the gain from training on just a single synthesized version of the training data, e.g., 81.4 $\rightarrow$ 88.0 F1 vs. 88.0 $\rightarrow$ 89.0 F1. \cref{fig:vary_synthetic_examples} shows how F1 varies as the number of synthetic copies of the training data is increased for a range of models.

While LID models trained on Dakshina pair \ngram model derived training data tend to perform well, irrespective of the order of the pair \ngram model, the samples from the Aksharantar-trained pair \ngram models are strictly worse. This gap persists even when also training on the original \BhashaAbhijnaanam training set. While the 1-best candidate generated from these pair \ngram models are complementary to the released training set, by themselves, the 1-best candidate can be quite poor (\cref{fig:synthetic_1best}). For example, training only on samples from the 2-gram pair LM yields 72.5 F1, far worse than the baseline of 81.4. But combining that data with the baseline training set yields strong improvements.

\section{Varying the number of harvested training examples}
\label{app:adding_natural_examples}


\begin{figure}[t]
    \centering
    \includegraphics[width=\linewidth]{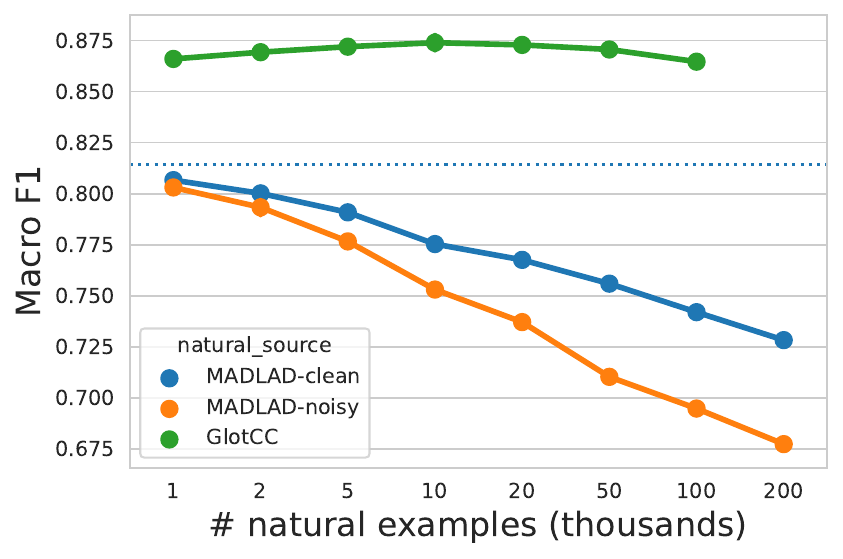}\vspace*{-0.05in}
    \caption{\footnotesize Macro development set F1 as a function of training corpus and number of examples added to each class. The original \BhashaAbhijnaanam training set is included in all runs. The dashed line corresponds to the performance of training only on the original \BhashaAbhijnaanam synthetic training set.}
    \label{fig:vary_natural_examples}\vspace*{-0.15in}
\end{figure}

\cref{fig:vary_natural_examples} presents development set F1 as the number of sampled examples added to the \BhashaAbhijnaanam training set is varied in $\{1, 2, 5, 10, 20, 50, 100, 200\}$ thousand examples. A fastText model was retrained in each case, and evaluated on the development set. We replicated each of these runs 5 times, averaging performance over the runs. Note that the original training set contains 100k examples per class, so adding up to 1k examples per class resulted in at most a 1\% increase in the training set size.

While the addition of more MADLAD data actively hurt the performance, the addition of the ``noisy'' subset harms the LID model more than the ``clean'' subset as the number of added examples increases. GlotCC is shown to be useful, even though this data also contains noisy labels, and does not cover all of the Dakshina classes.  See \cref{tab:glotcc_examples} for some illustrative examples from the GlotCC dataset. Label quality is paramount in training romanized LID models. Harvested text has little value, or can even be actively harmful, if it is unlikely to actually be the language of interest.

\section{Spelling variation in synthetic romanized samples}
\label{app:synthetic_spelling_variation}

We find that training on synthetic samples of romanized text improves LID classification performance. Do these synthetic samples actually reflect natural spelling variation? In this section, we compare the variation resulting from sampling versus 1-best decoding from the pair 3-gram transliteration models trained on Dakshina romanization lexicons. Because each decoded/sampled token was derived from the same native script string, we are able to directly compare spelling variation between romanized words generated by each method, token-by-token.

Over all Dakshina languages, 31\% of synthetically romanized tokens differ between the 1-best decoded and sampled romanizations. To find common patterns of these differences, we collected counts of character 5-gram minimum Levenshtein distance edits between the 1-best decode vs. sampled tokens within each language's synthetic training set. For example, we counted all instances where a substring ``arne'' in the 1-best romanization was sampled as ``arane'' by the FST romanization model, or ``attu'' was sampled as ``atthu''. From these counts we can identify common surface edit patterns.

\begin{table*}[t]
\centering%
\begin{adjustbox}{width=0.9\textwidth}
\begin{tabular}{clp{13cm}}
\toprule
Label & Count (\%) & Description \\
\midrule
\textbf{A} & 9  (4.5) & Change a low/mid vowel (\textbf{A}/e/o). Indicative of variation in pronunciation. \\
\textbf{G} & 14 (7.0) & Inserting an additional consonant at the same place of articulation. Indicative of \textbf{G}emination. \\
\textbf{H} & 19 (9.5) & Addition of \textbf{H} following a consonant. Indicative of aspiration. \\
\textbf{I} & 51 (25.5) & Inserting/deleting a vowel after a consonant. Indicative of \textbf{I}mplicit vowel representation. \\
\textbf{L} & 98 (49.0) & Addition/deletion of a vowel next to existing vowel. Indicative of vowel \textbf{L}ength/quality. \\
\textbf{N} & 2 (1.0) & Addition of syllable-final \textbf{N}asal. Indicative of variation in representing \emph{anusvara}. \\
\textbf{V} & 5 (2.5) & Consonant change at same place of articulation. Indicative of alternation in \textbf{V}oicing \\
\bottomrule
\end{tabular}
\end{adjustbox}
\caption{Description of spelling variation labels in \cref{tab:top20_spelling_variations}. Counts are over a total of 200 5-gram edit patterns, the 20 most frequent per language. Counts do not sum to 200 as one edit pattern was labeled as both {\bf A} and {\bf L}, and three patterns fell outside of this coding scheme.}
\label{tab:spelling_variation_key}
\end{table*}

\begin{table*}[t]
\begin{small}
\begin{tabular}{@{}c@{~~~}c@{~~~}c@{~~~}c@{~}|c@{~~~}c@{~~~}c@{~~~}c@{~}|c@{~~~}c@{~~~}c@{~~~}c@{~}|c@{~~~}c@{~~~}c@{~~~}c@{}}\toprule
Cnt x& 1-best & Sample & La- & Cnt x& 1-best & Sample & La- & Cnt x& 1-best & Sample & La- & Cnt x& 1-best & Sample & La- \\
1000 & \ngram & \ngram & bel & 1000 & \ngram & \ngram & bel & 1000 & \ngram & \ngram & bel & 1000 & \ngram & \ngram & bel\\\midrule
\multicolumn{4}{c|}{\bf Bangla} & \multicolumn{4}{c|}{\bf Gujarati} & \multicolumn{4}{c|}{\bf Hindi} & \multicolumn{4}{c}{\bf Kannada}\\
4.0 & yeech & yech & L & 3.4 & maate & mate & L & 2.4 & karn & karan & I & 3.2 & avag & avaag & L \\
3.6 & eeche & eche & L & 3.2 & karv & karav & I & 2.1 & kart & karat & I & 2.6 & vagi & vaagi & L \\
3.3 & ebong & abong & A & 2.8 & vama & vaman & M & 1.7 & arne & arane & I & 2.3 & hara & haara & L \\
3.1 & hayee & haye & L & 2.8 & kari & karee & L & 1.6 & rajya & rajy & I & 2.2 & kara & kaara & L \\
2.5 & ayeec & ayec & L & 2.3 & arva & arava & I & 1.5 & yojan & yojn & I & 2.0 & havaa & hava & L \\
2.4 & hayee & hoye & AL & 2.2 & yare & yaare & L & 1.5 & kuch & kuchh & H & 1.9 & thava & thav & I \\
2.0 & ebong & ebang & A & 2.1 & athi & athee & L & 1.4 & ojana & ojna & I & 1.9 & alag & alaag & L \\
1.9 & samy & samay & I & 1.9 & hati & hatee & L & 1.3 & hetr & hetra & I & 1.8 & matt & matth & H \\
1.7 & koren & karen & A & 1.8 & arva & arvaa & L & 1.2 & sakt & sakat & I & 1.7 & agid & aagid & L \\
1.6 & kore & korey & L & 1.8 & aman & amaan & L & 1.2 & dwara & dvara & !W & 1.5 & akar & akaar & L \\
1.6 & heke & hekey & L & 1.8 & rite & reete & L & 1.2 & lakin & lekin & A & 1.5 & attu & atthu & H \\
1.6 & bosth & basth & A & 1.8 & thay & thaay & L & 1.1 & bhara & bhar & I & 1.4 & tara & thara & H \\
1.6 & ayeec & oyec & L & 1.8 & hata & hataa & L & 1.1 & harat & hart & I & 1.4 & iyag & iyaag & L \\
1.5 & eche & echey & L & 1.7 & dhar & dhaar & L & 1.1 & karya & kary & I & 1.3 & haagu & hagu & L \\
1.5 & proti & prati & A & 1.6 & hava & havaa & L & 1.1 & arte & arate & I & 1.3 & aman & amaan & L \\
1.5 & chil & chhil & H & 1.6 & vama & avama & I & 1.1 & arti & arati & I & 1.3 & aagu & aagoo & L \\
1.5 & korec & karec & A & 1.6 & aara & aaraa & L & 1.1 & nahin & nahi & M & 1.3 & haag & haago & I \\
1.4 & eche & echhe & H & 1.6 & vama & vamaa & L & 1.0 & karan & karn & I & 1.3 & anta & antha & H \\
1.4 & tini & teeni & L & 1.5 & adha & adhaa & L & 1.0 & tarh & tarah & I & 1.3 & mana & maana & L \\
1.4 & korte & karte & A & 1.5 & tyar & tyaar & L & 0.9 & pahl & pahal & I & 1.3 & kari & kaari & L \\ \midrule
\multicolumn{4}{c|}{\bf Malayalam} & \multicolumn{4}{c|}{\bf Marathi} & \multicolumn{4}{c|}{\bf Punjabi} & \multicolumn{4}{c}{\bf Tamil}\\
3.8 & nathu & nath & H & 4.0 & karn & karan & I & 9.1 & icch & ichch & H & 8.4 & athth & ath & G \\
3.5 & amaay & amay & L & 2.4 & arny & arany & I & 9.1 & vicc & vichc & H & 7.4 & ththi & thi & G \\
2.4 & yaanu & yanu & L & 2.4 & rnya & ranya & I & 6.4 & vicch & vich & G & 5.9 & nth & nthth & G \\
2.4 & thaan & than & L & 2.3 & arata & arta & I & 3.3 & karan & karn & I & 5.4 & ththu & thu & G \\
2.3 & maayi & mayi & L & 1.8 & athi & athee & L & 2.3 & dian & diaan & L & 5.3 & argal & arkal & V \\
2.1 & haana & hana & L & 1.7 & karat & kart & I & 2.0 & keeta & kita & L & 5.3 & uthth & uth & G \\
1.9 & maaya & maya & L & 1.7 & asun & asoon & L & 2.0 & dian & diyan & L & 5.1 & ththa & tha & G \\
1.8 & aayir & ayir & L & 1.5 & achi & achee & L & 1.9 & jand & jaand & L & 4.6 & ithth & ith & G \\
1.8 & ikka & ikkaa & L & 1.5 & harat & hart & I & 1.6 & khia & khiaa & L & 4.4 & antha & andha & V \\
1.8 & athaa & atha & L & 1.4 & arna & arana & I & 1.5 & ahin & aheen & L & 4.2 & hthil & thil & G \\
1.8 & sthaa & stha & L & 1.4 & sath & sathe & I & 1.4 & hian & hiaan & L & 4.1 & aigal & aikal & V \\
1.7 & kkan & kkaan & L & 1.3 & adhi & adhee & L & 1.3 & keeti & kiti & L & 4.1 & ththa & ttha & H \\
1.6 & amaan & aman & L & 1.3 & bhara & bhar & I & 1.3 & anda & aanda & L & 4.0 & anga & angka & G \\
1.5 & aanam & anam & L & 1.3 & madhy & madh & !Y & 1.2 & nahi & nahee & L & 3.8 & runth & rundh & V \\
1.4 & hamaa & hama & L & 1.2 & adhye & adhe & !Y & 1.2 & bach & bacch & G & 3.7 & ngal & ngkal & G \\
1.4 & maanu & manu & L & 1.2 & arun & aroon & L & 1.0 & aria & ariaa & L & 3.7 & ithth & itth & H \\
1.3 & undaa & unda & L & 1.1 & nyat & anyat & I & 1.0 & ghat & ghatt & G & 3.5 & tha & ththa & G \\
1.3 & laanu & lanu & L & 1.1 & mhana & mhan & I & 1.0 & karan & kran & I & 3.5 & ththu & tthu & H \\
1.3 & thram & tram & H & 1.1 & arne & arane & I & 1.0 & vale & vaale & L & 3.4 & athth & atth & H \\
1.3 & ayaan & ayan & L & 1.1 & hoti & hotee & L & 0.9 & arti & arati & I & 3.2 & galai & kalai & V \\ \midrule
\multicolumn{4}{c|}{\bf Telugu} & \multicolumn{4}{c|}{\bf Urdu}\\
4.2 & unna & unnaa & L & 16.0 & ahein & ahin & L \\
3.0 & nchaa & ncha & L & 14.5 & nahei & nahi & L \\
2.8 & incha & inch & I & 8.6 & allah & alla & H \\
2.7 & nnar & nnaar & L & 8.1 & ahein & ahen & L \\
2.7 & chaar & char & L & 7.4 & nahei & nahe & L \\
2.5 & amlo & amloo & L & 6.2 & sath & saath & L \\
2.3 & naru & naaru & L & 5.8 & stan & astan & I \\
2.2 & hara & haara & L & 5.3 & ksta & kasta & I \\
2.1 & anta & antha & H & 5.2 & akst & akast & I \\
2.0 & aalan & alan & L & 5.2 & paks & pakas & I \\
1.9 & tunn & tunna & I & 4.4 & rahe & rahay & L \\
1.9 & aanik & anik & L & 4.1 & waqat & waqt & I \\
1.8 & haaru & haru & L & 4.1 & khla & khala & I \\
1.7 & anik & aanik & L & 3.8 & stan & istan & I \\
1.7 & chaal & chal & L & 3.7 & arne & arnay & L \\
1.7 & aanni & anni & L & 3.6 & ksta & kista & I \\
1.6 & tara & thara & H & 3.6 & akst & akist & I \\
1.6 & nnay & nnaay & L & 3.6 & paks & pakis & I \\
1.6 & tana & thana & H & 3.5 & karn & karna & I \\
1.6 & dhaan & dhan & L & 3.3 & kart & karta & I \\
\end{tabular}
\end{small}
\caption{\footnotesize The twenty most frequent 5-gram spelling variations between 1-best and sampled romanized tokens per Dakshina language. Each instance is labeled according to the key in \cref{tab:spelling_variation_key}. The three exceptions noted in the text are denoted by ``!W'' and ``!Y'' labels. Counts (in thousands, \emph{Cnt x 1000}) are over the entire \BhashaAbhijnaanam training set within each language.}
\label{tab:top20_spelling_variations}
\end{table*}

How nasals, implicit vowels, aspiration, vowel quality, and voicing are rendered in romanizations can vary in a language – sometimes these phenomena are clearly indicated in the Latin script, sometimes not \cite{roark-etal-2020-processing, demirsahin-etal-2022-criteria, kirov-etal-2024-context}.  We coded each of the 20 most frequent 5-gram edit patterns within a language by surface pattern type (\cref{tab:spelling_variation_key}). Each of these patterns are representative of natural romanization variation attested in the literature and we share the full set of annotated patterns in \cref{tab:top20_spelling_variations}.

Of these surface patterns, inserting vowels next to an existing one to indicate quality or length (label \textbf{L}, 49\%) was the most common. This includes alternation between ``u'' $\leftrightarrow$ ``oo'' and ``i'' $\leftrightarrow$ ``ee'' to indicate IPA /u/, /i/ respectively, along with doubling of vowels to explicitly indicate vowel quality (e.g., to distinguish from \emph{schwa}), ``a'' $\leftrightarrow$ ``aa'', ``e'' $\leftrightarrow$ ``ee'', and ``i'' $\leftrightarrow$ ``iy''.

The second most common pattern (label \textbf{I}, 25.5\% of patterns) was addition/deletion of a lone vowel following a consonant. This is indicative of whether the implicit vowel (e.g., schwa) was explicitly written in the romanization. For the surface patterns we see here, this vowel is ``a'' (e.g., in Hindi ``yojan'' $\rightarrow$ ``yojn'').

The addition of an ``h'' following a consonant often indicates aspiration (label \textbf{H}, 9.5\%), but may also clarify some other property of the consonant. For instance, in Tamil, ``th'' and ``dh'' respectively indicate unvoiced and voiced \emph{dental} stops -- the voicing property depends on the context and is not indicated in native Tamil orthography~\cite{schiffman:1999,annamalai:2015}. Both of these correspond to unaspirated consonants, where the presence of ``h'' instead distinguishes them from their retroflex counterparts.

The final most frequent pattern was doubling of a consonant at the same place of articulation, e.g., ``tha'' $\rightarrow$ ``ththa'' (label \textbf{G}, 7\% of patterns). This pattern occurred frequently in Tamil and Punjabi edit strings. In Tamil, a Dravidian language with agglutinative morphology, gemination is often due to the \emph{sandhi} effect, a set of phonological changes occurring at the location where morphemes combine~\cite{ciotti:2017}. Gemination is typical for Punjabi as well, unlike other Indo-Aryan languages \cite{bhatia:2010}.

Other patterns occurring in less than 5\% of edits were alternation in the choice of a back/mid vowel (e.g., frequent a/o alternation in Bangla), changing the voicing of a consonant perhaps due to coarticulation effects with a neighboring vowel (e.g., ``aigal'' $\rightarrow$ ``aikal''), and the addition/deletion of a syllable-final nasal due to transliterating the \emph{anusvara} character explicitly (``nahin'' $\rightarrow$ ``nahi'').

Out of these 200 frequent 5-gram edit patterns, we were only unable to classify three instances as one of the given classes. Of these, ``dwara'' $\rightarrow$ ``dvara'' in Hindi indicates a plausible variation, as v/w are allophones of each other in Hindustani \cite{pierrehumbert:1996}. In Marathi, ``madhy'' $\rightarrow$ ``madh'' and ``adhye'' $\rightarrow$ ``adhe'' indicates Latin spelling variation in a common morpheme, \includegraphics[height=0.3cm]{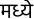}, meaning ``in'' or ``amid''.





\section{Evaluation on a synthetic development set}
\label{app:dakshina_dev_motivation}

We chose to develop LID models on natural romanized examples from the Dakshina development set, restricting ourselves to the subset of Dakshina languages where we had natural examples to evaluate on. \cref{tab:ba_synthetic_dev_perf} displays the performance of these models on the synthetic development set for the same subset of languages.

We find that model performance is higher than what we found in \cref{tab:sample_results,tab:union_results,tab:copies_results,tab:harvest_results}. This agrees with the observation in Table 5 of \citet{madhani-etal-2023-bhasa} that LID performance on automatically transliterated data is clearly inflated over naturally-generated text. The variation in performance across different training sets is also narrowed, potentially making training set selection more difficult: between 82.2\% and 90.5\% accuracy for the natural development set vs. 87.2\% to 93.1\% for the synthetic.


\begin{table*}[t]
    \centering%
\small%
\begin{tabular}{l|rr|rr}
\toprule
\multirow{3}{*}{Included training data} & \multicolumn{2}{c|}{Not Including B-A} & \multicolumn{2}{c}{Including B-A}\\\cmidrule{2-5}
 & fastText & mT5-large & fastText & mT5-large \\
 & Acc / F1 & Acc / F1  & Acc / F1 & Acc / F1 \\
\midrule
None &  && 87.2 / 86.8 & 89.3 / 89.2 \\
\midrule
 Dakshina 3-gram 1x samples & 90.5 / 90.3 & 89.4 / 89.3 & 92.1 / 92.0 & 91.8 / 91.7 \\
 Dakshina 3-gram 10x samples & {\bf 91.4 / 91.3} & 89.7 / 89.7 & 92.4 / 92.3 & 91.4 / 91.3 \\
 Dakshina 2-gram & 88.8 / 88.6 & 88.8 / 88.7 & 91.8 / 91.6 & 90.9 / 90.8 \\
 Dakshina 4-gram & 91.1 / 90.9 & {\bf 90.0 / 90.0} & 92.0 / 91.9 & 91.2 / 91.2 \\
 Aksharantar 3-gram & 89.2 / 88.8 & 88.8 / 88.4 & 90.9 / 90.7 & 89.8 / 89.5 \\
 Aksharantar 3-gram 10x samples & 90.5 / 90.3 & 89.3 / 89.0 & 91.4 / 91.2 & 89.9 / 89.6 \\
GlotCC & &&90.7 / 90.6 & 91.4 / 83.8 \\
Dakshina 3-gram 10x + GlotCC & && {\bf 93.1 / 93.2} & {\bf 92.4 / 92.4} \\
\bottomrule
\end{tabular}
\normalsize%
\vspace*{-0.05in}
\caption{\footnotesize Model performance on the synthetic development set released with the B-A corpus, for languages included in Dakshina.}
\label{tab:ba_synthetic_dev_perf}\vspace*{-0.15in}
\end{table*}

\section{Error analysis} \label{app:errors}
Comparing individual development set predictions from the baseline fastText model trained on \BhashaAbhijnaanam synthetic training data with the model trained on the best performing synthetic training data (\BhashaAbhijnaanam training data combined with Dakshina 3-gram sampled x 10 training data), we find that out of the 11,000 examples in the development set: (1) 8,847 examples were correctly classified by both models; (2) 163 were regressions from the baseline; (3) 989 were newly classified correctly by the updated model; and (4) 1,001 examples were incorrectly classified by both models. 
It is worth noting that the examples that were incorrectly classified by the updated model are markedly shorter than the ones which were correctly classified -- median of 3 tokens and 20 characters for the ``both lose'' case, 12 tokens and 90 characters for the ``both win'', 4 tokens and 29 characters for the ``regression'' case, and 11 tokens and 64 characters for the ``corrected'' case. \cref{tab:dakshina_dev_random_errors} includes some examples randomly selected from each group. Many of these incorrect examples contain English strings, either wholly or in part,\footnote{Note that the \BhashaAbhijnaanam benchmark removed many such items from their test set, so this is one difference between our development and test sets.} and proper names are also common (possibly confusing the LID model).


\begin{table*}[ht!]
\centering%
\begin{adjustbox}{width=\linewidth}
\small%
\begin{tabular}{l|p{11.5cm}}
\toprule
Language & Examples \\ \midrule
Bangla & ``ALLAH AMR ROB, NOBI AMR SOB. ISLAM AMR DHORMO, NAMAZ UTTOM KORMO.'', ``aar aami dekhlam, maar chokh duto anande nachchhe.'', ``usher alter'', ``This story was co-written by a member of our community using our AI powered storyteller.'' \\ \midrule
Gujarati & ``Happy Holi quotes and status'', ``Himalaya Rudraksh \& Gems Testing Lab - India's Most Trusted Rudraksha, Diamond \& Gemstone Testing Laboratory'', ``Rasulullah Syed al Mursalin ane Khaatam al Nabiyyin chhe.'' \\ \midrule
Hindi & , ``Shamooael 30:1 teesare din jab daud apane janon samet sikalag pahuncha, tab unhon ne kya dekha, ki amalekiyon ne daakkhian desh aur sikalag par chaddhai kee. aur sikalag ko mar ke foonk diya, 6.'', ``Natural Ways to Improve Memory in Hindi: Yaaddasht Badhaye'', ``Cricket satta ka vikas ek aise vyavsay ki or ishara karta hai jo samay ke saath badalta hai.'' \\ \midrule
Sindhi & ``room aeron chari room aeron chai rroom aeron chairr oom aeron chair orom aeron chair room aeron chair romo qeronchairroom weronchairroom seronchairroom xeronchairroom zeronchairroom eeronchairroom aaeronchairroom eronchairroom a2ronchairroom a3ronchairroom a4ronchairroom awronchairroom arronchairroom asronchairroom adronchairroom afronchairroom aaronchairroom aeeronchairroom...'', ``If you like to book room in a Aeron chair room use ````Check price and availability or'' ````Book now'''' green button, then you will be redirected to the main booking site from our partners, where you would select date of booking and check prices and availability of hotel rooms.'', ``Superstar Ayeza Khan touches new skies of popularity by performing in hit dramas “Chupke Chupke” and “Mere Pass Tum Ho”.'' \\ \midrule
Tamil & ``appoathu moayeesan avarka'lai noakki: kadavu'lin aaseer ungka'lukkuk kidaikkumpadi in'ru ungka'lil ovvoruvanum than than makanaiyum sakoatharanaiyum pazhivaangkinamaiyaal, aa'ndavarukku ungka'l kaika'lai arppa'nam seytheerka'l en'raar.'', ``paaravoanum avan oozhiyarka'l anaivarum ekipthiyar yaavarum iravil ezhunthanar.'', ``Naan kankalai mooti thoonguvadhu pol natiththen.'' \\ \midrule
Telugu & ``Ela undi ani adiga chala bagundi eppudu ela enjoy cheyaledu ani hug chesukundi night 12 ayyindi elago evaru leru chuttu koncham rest tesukoni tent pakkana plana chesam pakkana fire undi'', ``Prati samvatsaram January 1 na Global family Day jarupukuntaru .'', ``Ee Nagaraniki Emaindi Meme Movie: We Arranged The Entire Movie In Meme Templates.'' \\ \midrule
Urdu & ``Koi khuwahish nahi is deewanay ki'', ``Tamaam hamd us Allah ke liye hai jo chupi hui cheezoun ki gehraaioun mein utra hua hai.'', ``Click here \texttt{https://youtu.be/iLyCJGOU7Js?si=jHKBes\_6T9DjYw0s}'' \\
\bottomrule
\end{tabular}
\normalsize%
\end{adjustbox}
\caption{Romanized sentences from a selection of development set languages in the GlotCC corpus. While some strings are plausibly the correct language, some English strings and boilerplate are included.}
\label{tab:glotcc_examples}
\end{table*}

\begin{table*}[t]
\centering%
\begin{adjustbox}{width=0.91\textwidth}
\begin{tabular}{@{}l@{~}c@{~}c@{~}c@{~}p{9cm}@{}r@{~}r@{}}
\toprule
& Gold & Baseline & Updated & & \multicolumn{2}{c}{Number of}\\
Bucket & Label & System & (Best) & Example & Tokens & Chars \\
\midrule
both win & Telugu & Telugu & Telugu & bhakti paaravasyamu saranaagati ivi ee aaluvaarula jeevitamlonoo rachanalalonoo vaarini gurinchina gaathalalonoo pramukhangaa kaanavachche amsaalu & 14 & 146 \\
both win & Punjabi & Punjabi & Punjabi & buneyadi adhikar manukhi azadi da muddla sidhant han ate harek bharti di shakhsiyat de sahi vikaas layi eh jaruri han & 20 & 117 \\
both win & Telugu & Telugu & Telugu & dharinchagaligina computer & 2 & 26 \\
both win & Bangla & Bangla & Bangla & bivinno arthonoitik totto o mukto bazar sunirdishto boishisto aaboshoyk udahoronswarup ekti nikhut bazar shathe nirbhul tottho ebong nikhut protijogitaar & 19 & 153 \\
both win & Bangla & Bangla & Bangla & gobar goho & 2 & 10 \\
both win & Punjabi & Punjabi & Punjabi & parkash kol Unni R ki kahani nu film layi lain di ek surantar yojna vi si & 16 & 73 \\
both win & Malayalam & Malayalam & Malayalam & ennaal netveyarinethire oru velluviliyuyarthaan ithinaayilla & 5 & 60 \\
both win & Tamil & Tamil & Tamil & Jamui makkalavaith thoguthi Inthiya makkalavaikkaana thoguthiyaagum Ithu Biharin 40 makkalavaith thoguthigalil ondru & 12 & 116 \\
both win & Hindi & Hindi & Hindi & 2007 Uttar Pradesh vidhan sabha chunav men inhone Uttar Pradesh kr Merath jile ke Merath kaint vidhan sabha nirvachan shetr se Bhajpa ki aur se chunaav men bhag liya & 29 & 165 \\
both win & Punjabi & Punjabi & Punjabi & Paraguay vich hundi gair kanunni jungal vaadhi & 7 & 46 \\\midrule
regression & Telugu & Telugu & Kannada & paurushamme pongeraa & 2 & 20 \\
regression & Telugu & Telugu & Kannada & janaganamana & 1 & 12 \\
regression & Urdu & Urdu & Hindi & jinhein ham dekh kar jeete the Nasir & 7 & 36 \\
regression & Marathi & Marathi & Punjabi & Bahut din nacha bhetalo saubhadra & 5 & 33 \\
regression & Bangla & Bangla & Malayalam & Fellow of the Association for Computing Machinary 1994 & 8 & 54 \\
regression & Marathi & Marathi & Gujarati & Narahar kurundakar smruti sahity sammelan Nanded & 6 & 48 \\
regression & Urdu & Urdu & Hindi & 1819 mein venezuela aur granada ne mil kar jamhooriya banayi jis ka naam columbia rakha gaya & 16 & 92 \\
regression & Urdu & Urdu & Hindi & chataanein aur romaan & 3 & 21 \\
regression & Punjabi & Punjabi & Kannada & es da vikaas pracina russi bhasa valon hoeya & 8 & 44 \\
regression & Telugu & Telugu & Tamil & adi vasanta kaalam & 3 & 18 \\\midrule
corrected & Hindi & Bangla & Hindi & wo Royal socity of London ke nirvachit sadasy the & 9 & 49 \\
corrected & Hindi & Punjabi & Hindi & Unhe Lhasa se beijing tak jana tha lekin aisa sambhav nahi hone par Sangpo yani Brahmaputra ya Bhutan ke raste Bharat ane ke nidesh diye gaye the & 27 & 145 \\
corrected & Punjabi & Urdu & Punjabi & Ek jahaad pyar de layi & 5 & 22 \\
corrected & Hindi & Punjabi & Hindi & sohan rahi ke anusar geet vidha sahitya ki sabse kathin evam shresth vidha hai & 14 & 78 \\
corrected & Telugu & Gujarati & Telugu & aasale adiyaasalai nadi vesavi bratukaayene & 5 & 43 \\
corrected & Urdu & Punjabi & Urdu & qaumi parast rahnuma & 3 & 20 \\
corrected & Tamil & Malayalam & Tamil & Aatkalam Kanitham & 2 & 17 \\
corrected & Sindhi & Bangla & Sindhi & saawanu men ute jaa bhagea panhinjon menhon dunad gion hdrie pke te kadhi indaa ahin jite paani kona hondo ahe & 20 & 110 \\
corrected & Hindi & Urdu & Hindi & sardiyon mein yaha bhaari barfbaari hoti hai aur jheel bhi jam jaati hai & 13 & 72 \\
corrected & Tamil & Hindi & Tamil & paarampariya nel & 2 & 16 \\\midrule
both lose & Hindi & Malayalam & Sindhi & Cardinal & 1 & 8 \\
both lose & Hindi & Punjabi & Kannada & yogita bali & 2 & 11 \\
both lose & Gujarati & Malayalam & Telugu & ravishankar mahaaraaj & 2 & 21 \\
both lose & Hindi & Marathi & Marathi & Rachana Parulkar Ajebade Panwar & 4 & 31 \\
both lose & Urdu & Kannada & Kannada & kaala shahzada & 2 & 14 \\
both lose & Sindhi & Punjabi & Punjabi & sancho khalifa rashdeen 2 sancho wazir sahibha sancho khalifa rashideen sancho sahiba & 12 & 85 \\
both lose & Malayalam & Kannada & Kannada & narabali & 1 & 8 \\
both lose & Gujarati & Sindhi & Sindhi & tal & 1 & 3 \\
both lose & Urdu & Punjabi & Hindi & agar silsila bharfaj hota ho to silsile ki infiradi istilahaat laziman sifar ki taraf pahunchen gin & 16 & 99 \\
both lose & Tamil & Malayalam & Malayalam & saara thattil & 2 & 13 \\
\bottomrule
\end{tabular}
\end{adjustbox}
\caption{Sample of instances where the baseline/updated (best) synthetically trained models agree/differ.}
\label{tab:dakshina_dev_random_errors}
\end{table*}

\end{document}